\documentclass[10pt,twocolumn,letterpaper]{article}

\usepackage[pagenumbers]{cvpr} % To force page numbers, e.g. for an arXiv version

\usepackage{graphicx}
\usepackage{amsmath}
\usepackage{amssymb}
\usepackage{booktabs}
\usepackage{array}
\usepackage{pifont}
\usepackage{capt-of,etoolbox}
\usepackage{bbding}
\usepackage{algorithm}
\usepackage{algorithmic}
\usepackage{xcolor,colortbl}
\usepackage{hhline}
\usepackage{multirow}

\usepackage[pagebackref,breaklinks,colorlinks]{hyperref}

\usepackage[capitalize]{cleveref}
\crefname{section}{Sec.}{Secs.}
\Crefname{section}{Section}{Sections}
\Crefname{table}{Table}{Tables}
\crefname{table}{Tab.}{Tabs.}

\def\cvprPaperID{2658} % *** Enter the CVPR Paper ID here
\def\confName{CVPR}
\def\confYear{2022}

\usepackage{overpic}
\usepackage{enumitem} %< control spacing in itemize/enumerate/...
\usepackage{overpic} %< add raw math symbols to figures
\usepackage{color}
\definecolor{turquoise}{cmyk}{0.65,0,0.1,0.3}
\definecolor{purple}{rgb}{0.65,0,0.65}
\definecolor{dark_green}{rgb}{0, 0.5, 0}
\definecolor{orange}{rgb}{0.8, 0.6, 0.2}
\definecolor{red}{rgb}{0.8, 0.2, 0.2}
\definecolor{darkred}{rgb}{0.6, 0.1, 0.05}
\definecolor{blueish}{rgb}{0.0, 0.3, .6}
\definecolor{light_gray}{rgb}{0.7, 0.7, .7}
\definecolor{pink}{rgb}{1, 0, 1}
\definecolor{greyblue}{rgb}{0.25, 0.25, 1}

\usepackage{blindtext}

\renewcommand{\paragraph}[1]{\vspace{1em}\noindent\textbf{#1}.}

\begin{document}
\title{Behind the Speech: How Much 3D Face Information is Embedded in Voices?}
\title{Cross-Modal Perceptionist: Can Face Geometry be Gleaned from Voices?}

\author{Cho-Ying Wu, Chin-Cheng Hsu, Ulrich Neumann \\
University of Southern California\\
{\tt\small \{choyingw, chincheh, uneumann\}@usc.edu}
}
\maketitle
\begin{abstract}

This work digs into a root question in human perception: can face geometry be gleaned from one's voices? Previous works that study this question only adopt developments in image synthesis and convert voices into face images to show correlations, but working on the image domain unavoidably involves predicting attributes that voices cannot hint, including facial textures, hairstyles, and backgrounds. We instead investigate the ability to reconstruct 3D faces to concentrate on only geometry, which is much more physiologically grounded. We propose our analysis framework, Cross-Modal Perceptionist, under both supervised and unsupervised learning. First, we construct a dataset, Voxceleb-3D, which extends Voxceleb and includes paired voices and face meshes, making supervised learning possible. Second, we use a knowledge distillation mechanism to study whether face geometry can still be gleaned from voices without paired voices and 3D face data under limited availability of 3D face scans. We break down the core question into four parts and perform visual and numerical analyses as responses to the core question. Our findings echo those in physiology and neuroscience about the correlation between voices and facial structures. The work provides future human-centric cross-modal learning with explainable foundations. See our \href{https://choyingw.github.io/works/Voice2Mesh/index.html}{project page}.
\end{abstract}
\section{Introduction}
\label{sec:intro}

This work studies to what extent voice can hint face geometry motivated by recent studies on voice-face matching and cross-modal learning \cite{nagrani2018seeing, Wen_2021_CVPR, zheng2021adversarial}. 
Many physiological attributes are embedded in voices. For example, speech is produced by articulatory structures, such as vocal folds, facial muscles, and facial skeletons, which are all densely connected.
Such a fact intuitively indicates potential correlations between voices and face shapes \cite{harrington2010acoustic}. 
Experiments in cognitive science point out that audio cues are associated with visual cues in human perception-- especially in recognizing a person's identity \cite{belin2004thinking}. 
Recent neuroscience research further shows that two parallel processing of low-level auditory and visual cues are integrated in the cortex, where voice processing affects facial structural analysis for the perception purpose \cite{young2020face}.

Traditional research in the voice domain focuses on utilizing voice inputs for predicting more conspicuous attributes which include speaker identity \cite{bull1983voice, maguinness2018understanding, ravanelli2018speaker}, age \cite{ptacek1966age, singh2016relationship, grzybowska2016speaker}, gender \cite{li2019improving}, and emotion \cite{wang2017learning, zhang2019attention}. 
A novel direction in recent development goes beyond predicting these attributes and tries to reconstruct \textit{2D face images from voice} \cite{NEURIPS2019_eb9fc349, oh2019speech2face, choi2020inference}. Their research is built on an observation that one can approximately envision how an unknown speaker looks when listening to the speaker's voice. Attempts towards validating this assumptive observation include the work \cite{oh2019speech2face} for image reconstruction and works \cite{NEURIPS2019_eb9fc349, choi2020inference} using generative adversarial networks (GANs). They aim to output face images from only a speaker's voice.

\begin{figure}[t!]
\begin{center}
\includegraphics[width=1.0\linewidth]{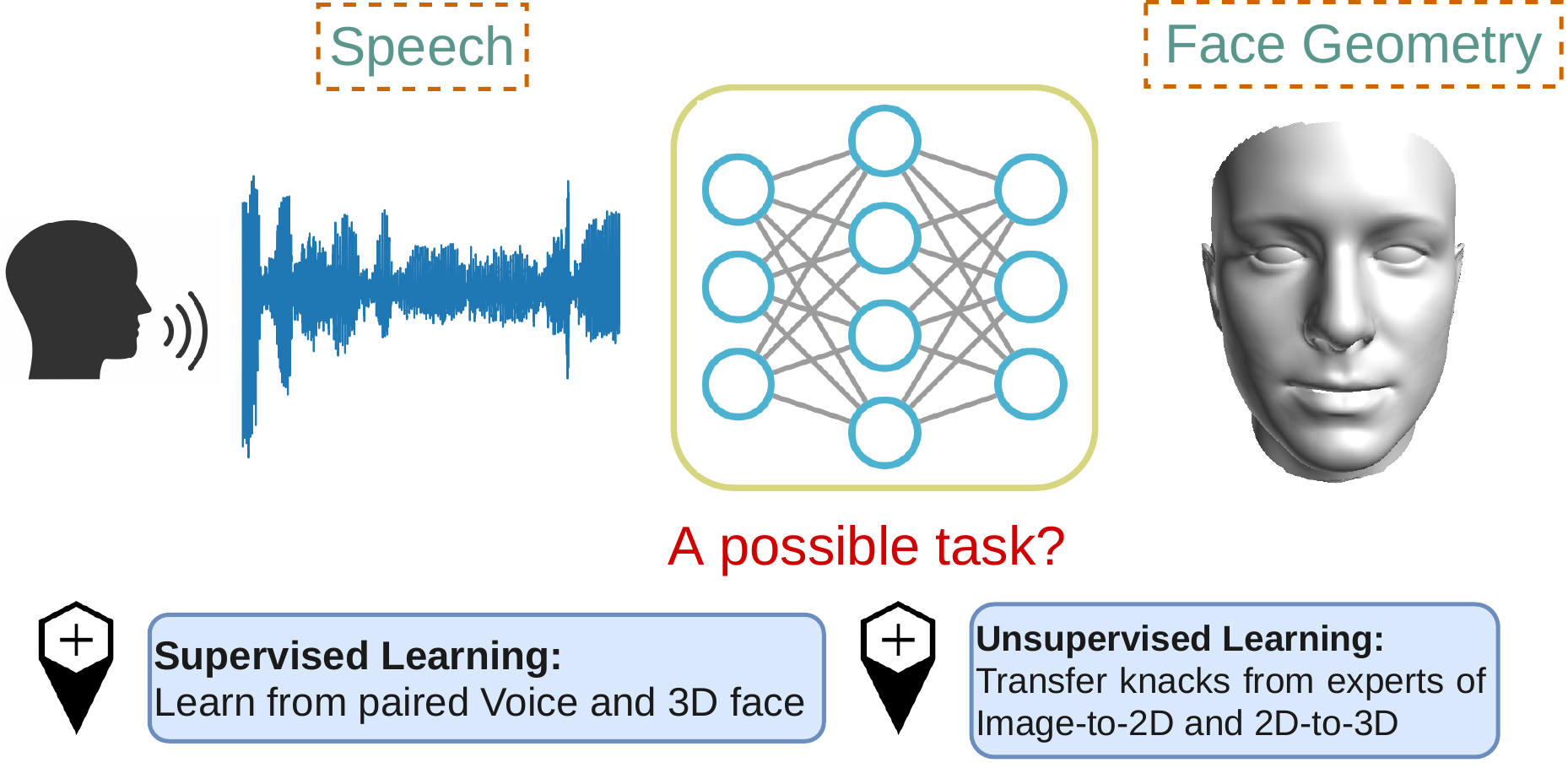}
\end{center}
  \vspace{-12pt}
  \caption{\textbf{Cross-Modal Perceptionist.} We study the correlations between voices and face geometry under both supervised and unsupervised learning settings. This work targets at more explainable human-centric cross-modal learning for biometric applications.}
\label{purpose}
\vspace{-6pt}
\end{figure} 

However, face images from voices are inherently ill-posed: the task involves predicting extraneous attributes that voices cannot hint, including image backgrounds, hairstyles, headgears, or beards. These attributes are apparently that one can choose without changing voices. 
Similar concerns arise regarding the correlations between voices and facial textures or ethnicity.
\cite{oh2019speech2face} demonstrates a t-SNE plot in which ethnicity is scattered across all samples, indicating its low correlations to voices.
As a result, quantifying the differences between an output face image and a reference is hard and less grounded.

Instead of producing face images, our analysis moves to the 3D domain with mesh representations and \textbf{predicts one's face geometry or skull structures from voices}, which is free from the above issues.
Working on 3D meshes is less ambiguous than images because the former includes less noisy variations unrelated to a speaker's voice, such as stylistic variations, hairstyles, background, and facial textures.
Moreover, meshes enable more straightforward quantification of differences between prediction and groundtruth in the Euclidean space-- unlike the case in using face images, where sources of differences involve backgrounds and hairstyles.

From the perspective of 3D faces, much research attention has been paid to 3D reconstruction from monocular images \cite{shang2020self,guo2020towards,zhu2016face, wu2021synergy} or video sequences \cite{garrido2016reconstruction, kim2018deep} for 3D face animation or talking face synthesis. In contrast, we are the first to investigate the correlations between one's 3D face geometry and voices, and we focus on the analysis of the face geometry gleaned from one's voices.
Our goal is to validate the correlations between voices and face geometry towards more explainable human-centric cross-modal learning with neuroscience support.

The analysis inevitably involves acquiring large-scale 3D face scans with paired voices, which is expensive and subject to privacy. To deal with this issue, we propose a novel \textit{Voxceleb-3D} dataset that includes paired voices and 3D face models.
Voxceleb-3D is inherited from two widely used datasets: Voxceleb \cite{nagrani2017voxceleb}) and VGGFace \cite{BMVC2015_41}, which include voice and face images of celebrities, respectively.
The approach \cite{Zhu_2015_CVPR} we adopt to create Voxceleb-3D is inspired by 300W-LP-3D \cite{zhu2016face}, the most-used 3D face dataset, and we will describe details in Sec.\ref{sec:methods/supervised}.

Our analysis framework \textbf{Cross-Modal Perceptionist} (CMP), investigates the feasibility to predict face meshes using 3D Morphable Models (3DMM, Sec.\ref{sec:methods/3dmm}) from voices on the following two scenarios (Fig. \ref{purpose}). 
We first train neural networks directly from Voxceleb-3D in a \textit{supervised learning} manner using the paired voices and 3DMM parameters (Sec.\ref{sec:methods/supervised}).
We further investigate an \textit{unsupervised learning} setting to inspect whether face geometry can still be gleaned without paired voices and 3D faces, which is a more realistic scenario. In this case, we use \textit{knowledge distillation} (KD) \cite{hinton2015distilling} to transfer knacks from the state-of-the-art method for 3D faces from images, SynergyNet \cite{wu2021synergy}, into our student network and jointly train speech-to-image and image-to-3D blocks (Sec.\ref{sec:methods/unsupervised}).

We design a set of metrics to measure the geometric fitness based on points, lines, and regions for both the supervised and the unsupervised scenarios.
The evaluation attempts to show correlations between 3D faces and voices with straightforward neural network-based approaches. The analysis with CMP enables us to comprehend the correlations between face geometry and voices. Our research lays explainable foundations for human-centric cross-modal learning and biometric applications using voice-face correlations, such as security and surveillance when only voice is given.

Our goal is not to recover high-quality 3D face meshes from voices comparable to synthesis from visual modalities such as image or video inputs, but we try to answer the core question under our CMP framework: can face geometry be gleaned from voice? We break down the question into four parts and will answer them through experiments.

Q1. Is it feasible to predict visually reasonable face meshes from voice?

Q2. How stable is the mesh prediction from different utterances of the same person?

Q3. Compared with face meshes produced by cascading separately trained speech-to-image and image-to-3D-face methods, can the performance of a joint training flow, where mesh prediction is trained with voice information, improve? How much?

Q4. What is the major improvement that voice information can bring in the joint training flow?

Our contributions are summarized.
\begin{enumerate}
  \item Towards explainable human-centric cross-modal learning, we are the first to study the correlations between face geometry and voices.
  \item We devise an analysis framework, Cross-Modal Perceptionist, which studies both supervised and unsupervised approaches to learn face meshes from voices. 
  \item We show extensive analysis and discussion and answer to four breakdown questions to validate the correlations between voices and face shapes
\end{enumerate}

%study whether face geometry can still be gleaned without face scans.
\section{Related Work}

%We document all related fields and methods for our framework into following sections.

\subsection{Audio: Learning Personal Traits from Voice}

The human voice is embedded with a wide range of personal information and has long been exploited for recognizing personal traits, 
such as speaker identity \cite{bull1983voice, maguinness2018understanding, ravanelli2018speaker}, age \cite{ptacek1966age, singh2016relationship, grzybowska2016speaker}, gender \cite{li2019improving}, and emotion status \cite{wang2017learning, zhang2019attention}.
Voices can also be used to monitor health conditions \cite{ali2017automatic} or applied to other medical applications \cite{han2021exploring}. 
Most existing works focus on predicting personal traits that are more intuitively related to voice. Our work can be seen as a much more challenging task for learning implicit personal faces or skull structures from voices. 

\subsection{Visual: 2D/3D Face Synthesis} % from Visual Modality}

Face-related synthesis has been under much research in the past years.
Generating 2D face images using GANs \cite{goodfellow2014generative, abdal2019image2stylegan, nie2020semi, Karras_2020_CVPR, richardson2021encoding, karras2019style} has been a prevalent task, and recent progress includes more realistic synthesis with diverse styles. The task of face reenactment \cite{garrido2014automatic, nirkin2019fsgan, thies2016face2face} focuses on transferring facial features from a source to a target. 
Some works focus on the 3D domain: synthesizing 3D face models from monocular images \cite{zhu2016face, guo2020towards, tran2018nonlinear, wu2021synergy}, synthesizing 3D face motion from videos \cite{kim2018deep, garrido2016reconstruction} using 3DMM \cite{egger20203d, tewari2021learning}, or implicit fields \cite{Yenamandra_2021_CVPR}.

%3D face modeling from monocular images adopt 3DMM to synthesize fa

\subsection{Audio-Visual Learning}
\textbf{Cross-Modal Face Matching} \cite{nagrani2018seeing,kim2018learning, Wen_2021_CVPR,ning2021disentangled,zheng2021adversarial} covers tasks where voices are used as queries to retrieve faces or vice versa.
These tasks are inherently \textit{selection} problems in which the best fit of a voice-face pair from the dataset is desired.
Another similar task is cross-modal verification \cite{Nawaz_2021_CVPR, tao2020audio,sari2021multi} that tells whether input faces and voices belong to the same person, which is a simply  \textit{classification} problem for paired inputs. Our work solves its root question and \textit{explains} the success in voice-face matching or verification by verifying correlations between voices and face geometry.

\textbf{Talking face synthesis} targets at generating coherent and natural lip movements.
Some works drive template images \cite{jamaludin2019you, zhou2019talking, guo2021adnerf} or template face meshes \cite{cudeiro2019capture} to talk by speech inputs.
Some replace lip movements in a video with movements inferred from another video or speech \cite{chen2018lip, wiles2018x2face}. Their focuses are coherent lip movements and thus are different from our target at studying holistic facial structures.

\textbf{Voice to Face} is the closest task to our work. This task is introduced recently to synthesize face images from only voice inputs. \cite{NEURIPS2019_eb9fc349} and \cite{choi2020inference} adopt GANs to generate face images from audio clips. \cite{oh2019speech2face} uses an encoder-decoder structure to reconstruct face images. However, the disadvantages are that 2D representations contain many variations, such as hairstyles, beards, backgrounds, and facial textures irrelevant to facial geometry, or the correlations lack physiological support. Besides, face reconstruction errors can be ambiguous because two images of the same person can contain different hairstyles and backgrounds.

Our analysis framework circumvents the issues raised by 2D face representations. 3D face models do not contain hairstyles, backgrounds, or texture variations. Geometric representation of meshes enables us to analyze the correlations between voices and 3D shapes and further directly measure gains and errors in the Euclidean space. In this way, we can focus on face geometry gleaned from voices.
\section{Method}
\label{sec:methods}
Our goal is to analyze how a person's voice relates to one's face geometry in the 3D space. Thus, we learn 3D face meshes using 3D Morphable Models (3DMM) from input speech and analyze the correlations under supervised and unsupervised learning settings. The supervised setting learns the correlation from a paired voice and 3D face dataset. The unsupervised learning studies a realistic case when such paired dataset is not available, is it still possible to predict face geometry from voice?  

\subsection{3D Morphable Models (3DMM)}
\label{sec:methods/3dmm}
3DMM \cite{egger20203d} is a popular method for 3D face modeling using principal component analysis (PCA). By estimating the weights of basis matrices, a 3D face can be constructed. 
We can decompose a face into two components: the average face and face shape variation. 
That is, for a face $A$, 
\begin{equation}
\vspace{-4pt}
      A = \bar{A} + V\alpha,
\label{3dmm_modeling}
\vspace{-4pt}
\end{equation}
where $\bar{A} \in \mathbb{R}^{3N}$ is the average face with $N$ three-dimensional vertices, 
$V \in \mathbb{R}^{3N\times P}$ is a basis matrix for the face shape variation, 
$\alpha \in \mathbb{R}^{P}$ is the coefficients.
Note that we can reshape $A$ into $A_r \in \mathbb{R}^{3\times N}$, 
a matrix representation suitable for 3D rotation and translation.

We set $N=53490$ vertices following BFM \cite{paysan20093d}, a particular form of 3DMM.
Per the dimensionality of shape variation basis, we choose $P=50$ following SynergyNet \cite{wu2021synergy}, the state-of-the-art 3D face reconstruction methods from \textit{images} using BFM.
There are 12 additional pose parameters in SynergyNet used to align reconstructed 3D faces to its 2D image inputs: a rotation matrix $R \in \mathbb{R}^{3 \times 3}$ and a translation vector $t \in \mathbb{R}^{3}$, i.e., $A_p = RA_r+t$.
In our analysis, we only use these pose parameters for visualizing how well a predicted face mesh fits a 2D shape outline.

\subsection{Supervised Learning with Voice/Mesh Pairs}
\label{sec:methods/supervised}
 
\begin{figure}[bt!]
\begin{center}
\includegraphics[width=1.0\linewidth]{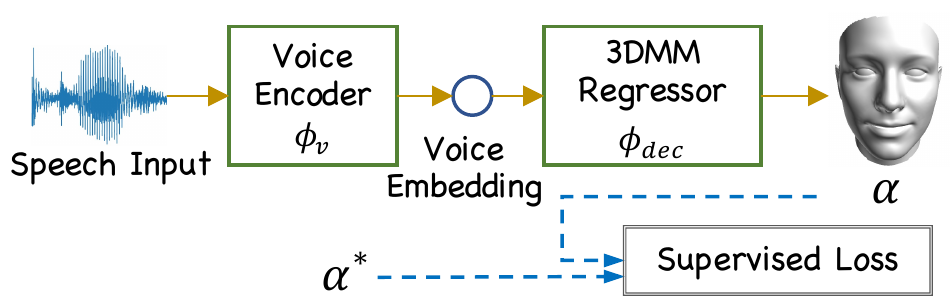}
\end{center}
  \vspace{-9pt}
  \caption{\textbf{Supervised learning framework.} Given a speech input, voice embedding is extracted by $\phi_v$. $\phi_{dec}$ then estimates 3DMM parameters $\alpha$ for 3D face modeling. The supervision is computed with groundtruth $\alpha^*$.}
\label{supervised_fig}
\vspace{-7pt}
\end{figure} 
 
We first describe the supervised learning setting, illustrated in Fig. \ref{supervised_fig}. Given a paired speech sequence and 3DMM parameters for an identity, we build an encoder-decoder structure first to extract voice embedding $v\in \mathbb{R}^{64}$ from a mel-spectrogram \cite{grochenig2001foundations}, which is a commonly used time-frequency representation for speech, of the input speech. Following \cite{NEURIPS2019_eb9fc349}, the voice encoder $\phi_v$ is pretrained on the large-scale speaker recognition task. Then, we train a decoder $\phi_{dec}$ to estimate 3DMM parameters, $\alpha$. We use groundtruth 3DMM parameters to supervise the training with $L_2$ loss.
\begin{equation}
      \mathcal{L}_{reg} = \|\alpha-\alpha^*\|^2
\label{supervised_loss}
\end{equation}
where $\alpha^*$ is groundtruth 3DMM parameters.

In addition, we adopt the triplet loss on the estimated 3DMM parameters $\alpha$. The triplet loss minimizes the difference of pairwise relations between (anchor, positive) and (anchor, negative) pairs with a soft margin.
\begin{equation}
\vspace{-3pt}
      \mathcal{L}_{tri} = max\{\|\alpha- \alpha_p\|_2-\|\alpha- \alpha_n\|_2+1,0\},
\label{triplet_loss}
\vspace{-3pt}
\end{equation}
where $\alpha$ plays as an anchor, $\alpha_p$ is a positive sample for the anchor, representing the same identity but regressed from different images, and $\alpha_n$, coming from a different identity, is a negative sample for the anchor. The triplet loss aims at coherent 3DMM parameters for the anchor and positive samples due to the same identities and simultaneously contrasting to the negative sample due to a different identity. The overall loss function is $\mathcal{L}_{sup} = \mathcal{L}_{reg}+\mathcal{L}_{tri}$.

The challenge of this supervised learning problem is how to obtain $\alpha^*$. Most large voice datasets, such as Voxceleb \cite{nagrani2017voxceleb}, only contain speech for celebrities, and most large face datasets, such as VGGFace \cite{BMVC2015_41}, only consist of publicly scraped face images. We first follow \cite{NEURIPS2019_eb9fc349} to fetch the intersection of voice and image data from Voxceleb and VGGFace. Then, we propose to fit 3D faces from 2D to create a novel dataset, \textbf{Voxceleb-3D}, using an optimization-based approach adopted by 300W-LP-3D \cite{zhu2016face}, the most-used 3D face dataset. In detail, we use an off-the-shelf 3D landmark detector \cite{bulat2017far} to extract facial landmarks from collected face images and then optimize 3DMM parameters to fit in the extracted landmarks. Our Voxceleb-3D contains paired voice and 3D face data to fulfill our supervised learning.

\subsection{Unsupervised Learning with KD}
\label{sec:methods/unsupervised}

Obtaining real 3D face scans is very expensive and limited by privacy, and the workaround of optimization-based 3DMM fitting with facial landmarks is time-consuming.
An unsupervised framework may serve real-world scenarios.
As a result, we propose an unsupervised framework with knowledge distillation. By leveraging a well-pretrained expert, it helps to validate whether face geometry can still be gleaned with neither real 3D face scans nor optimized 3DMM parameters.

%values of the unsupervised framework lie in studying how 3D faces can be reconstructed from voices without groundtruth 3D face scans or 3DMM parameters, which is closer to real-world scenarios.

\begin{figure}[bt!]
\begin{center}
\includegraphics[width=1.0\linewidth]{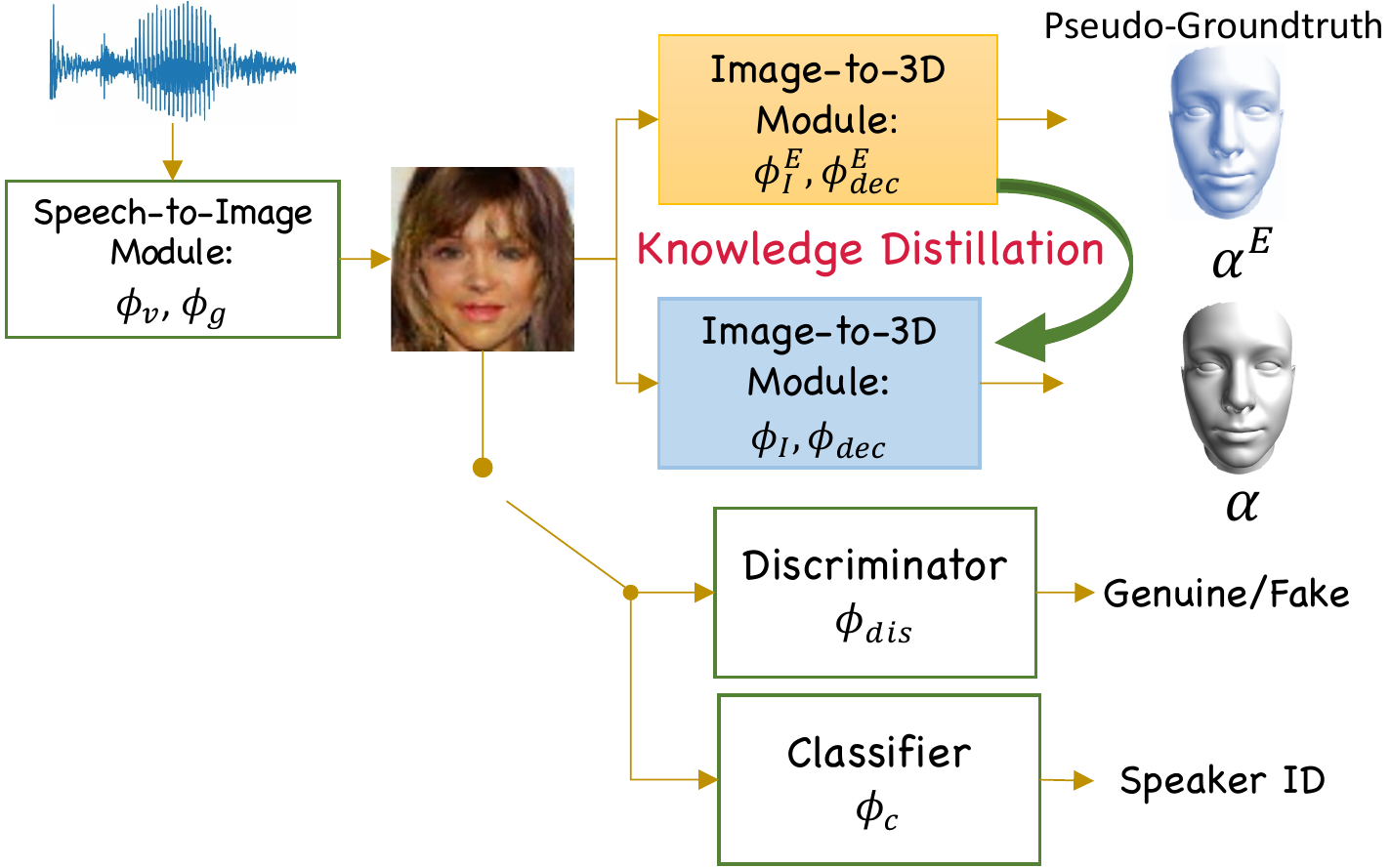}
\end{center}
  \vspace{-11pt}
  \caption{\textbf{Unsupervised learning with KD.} The unsupervised framework contains a GAN for face image synthesis with voice encoder $\phi_v$, generator $\phi_g$, discriminator $\phi_{dis}$, and classifier $\phi_c$. Then, knowledge distillation is used to achieve unsupervised learning, where information of image-to-3D-face mapping distilled from the expert network (yellow block) is exploited to train the student network (blue block). 2D face is a latent representation in this fashion. Beside using pseudo-groundtruth $\alpha^E$ to train the student, we also distill knowledge at intermediate layers using conditional probability distributions.}
  \vspace{-7pt}
\label{kd_pipeline}
\end{figure}

Our unsupervised framework, illustrated in Fig. \ref{kd_pipeline}, has two stages: (1) synthesizing 2D face images from voices with GAN and (2) 3D face modeling from synthesized face images. The motivation is that we first use the GAN to generate 2D faces from voices to obtain the speaker's appearance. However, 2D images contain variations of backgrounds, textures, hairstyles that are irrelevant to voice. Thus, the second-stage image-to-3D-face module disentangles geometry from other variations.

\textbf{Synthesizing face images from voices with GANs.}
Previous research develops a GAN-based speech-to-image framework \cite{NEURIPS2019_eb9fc349}. A voice encoder $\phi_v$ extracts voice embeddings from input speech. Then a generator $\phi_g$ synthesizes face images from the voice embeddings, and a discriminator $\phi_{dis}$ decides whether the synthesis is indistinguishable from a real face image. Last, a face classifier $\phi_c$ learns to predict the identity of an incoming face, ensuring that the generator produces face images that are truly close to the identity in interest. Here we overload notations of $\phi_v$ and other components introduced later for 3D face modeling in both Sec.\ref{sec:methods/supervised} and \ref{sec:methods/unsupervised} due to the same functionalities.

In detail, given a speech input $S$, its corresponding speaker ID $id$, and real face images $I_r$ for the speaker, the image synthesized from the generator is $I_f = \phi_g(\phi_v(S))$. The loss formulation is divided into two parts: real and fake images. For real images, the discriminator learns to assign them to "real" ($r$) and the classifier learns to assign them to $id$. The loss for real images is $\mathcal{L}_r = \mathcal{L}_d(\phi_{dis} (I_r), r)+\mathcal{L}_c(\phi_c (I_r), id)$ showing the discriminator and classifier losses respectively. For fake images, after producing $I_f$ from $\phi_g$, the discriminator learns to assign them to "fake" ($\bar{r}$) and the classifier also learns to assign them to $id$. The loss counterpart for fake images is $\mathcal{L}_f = \mathcal{L}_d(\phi_{dis} (I_f), \bar{r})+\mathcal{L}_c(\phi_c (I_f), id)$.

\textbf{3D face modeling from synthesized images.} After image synthesis by GAN, we build a network to estimate 3DMM parameters from fake images. The parameter estimation consists of an encoder $\phi_{I}$ and an decoder $\phi_{dec}$ to obtain 3DMM parameters $\alpha = \phi_{dec}(\phi_{I}(I_f))$. 3D face meshes are then reconstructed by Eq. \ref{3dmm_modeling}.

\textbf{Knowledge distillation for unsupervised learning}
\label{sec:unsuper}

To fulfill the unsupervised training, we distill the knowledge of image-to-3D-face reconstruction from a pretrained expert network. The expert, consisting of encoder $\phi_{I}^E$ and decoder $\phi_{dec}^E$, reconstructs 3D face models from synthesized face images and produces pseudo-groundtruth of 3DMM parameters $\alpha^E$. $\alpha^E$ is used to train the student network by $L_2$ loss:
\begin{equation}
     \mathcal{L}_{p-gt}=\|\alpha^E-\alpha\|^2. 
\label{loss_pgt}
\end{equation}
This KD strategy \textit{circumvents the needs of paired voice and 3D face data} and helps us achieve unsupervised learning.

In addition to pseudo-groundtruth, we also distill knowledge at intermediate layers and minimize their distribution divergence between the expert and the student. We measure the distributions in the feature spaces by the extracted image embedding $z^E \in \mathbb{R}^{B \times \nu}$ and $z \in \mathbb{R}^{B \times \nu}$ of the expert and the student network. We maintain the batch dimension $B$ and collapse the rest to $\nu$. Then as in \cite{passalis2020probabilistic}, we calculate the conditional probability $z$ between feature points as follows.
\begin{equation}
\footnotesize
     z_{i|j}=\frac{K(z_i,z_j)}{\sum_{k,k\ne j}K(z_k,z_j)},
     z^E_{i|j}=\frac{K(z_i^E,z_j^E)}{\sum_{k,k\ne j}K(z^E_k,z^E_j)},
\label{loss_pgt}
\end{equation}
where $K(\cdot, \cdot)$ is scaled and shifted cosine similarity whose outputs lie in [0,1]. Kullback-Leibler (KL) divergence is then used to minimize the two conditional distributions.
\begin{equation}
\footnotesize
     \mathcal{L}_{div}=\sum_i \sum_{j\ne i} z^E_{j|i} \text{log}\left( \frac{z^E_{j|i}}{z_{j|i}}\right). 
\label{loss_pgt}
\end{equation}
The KD loss is $\mathcal{L}_{\textit{KD}}=\mathcal{L}_{p-gt}+\mathcal{L}_{div}$. The overall unsupervised learning loss is combined with GAN loss and also triplet loss in Eq.\ref{triplet_loss}.
\begin{equation}
     \mathcal{L}_{unsuper}=\mathcal{L}_{f}+\mathcal{L}_{r}+\mathcal{L}_{\textit{KD}}+ \mathcal{L}_{tri}. 
\label{loss_pgt}
\vspace{-6pt}
\end{equation}
\begin{figure}[bt!]
\begin{center}
\includegraphics[width=1.0\linewidth]{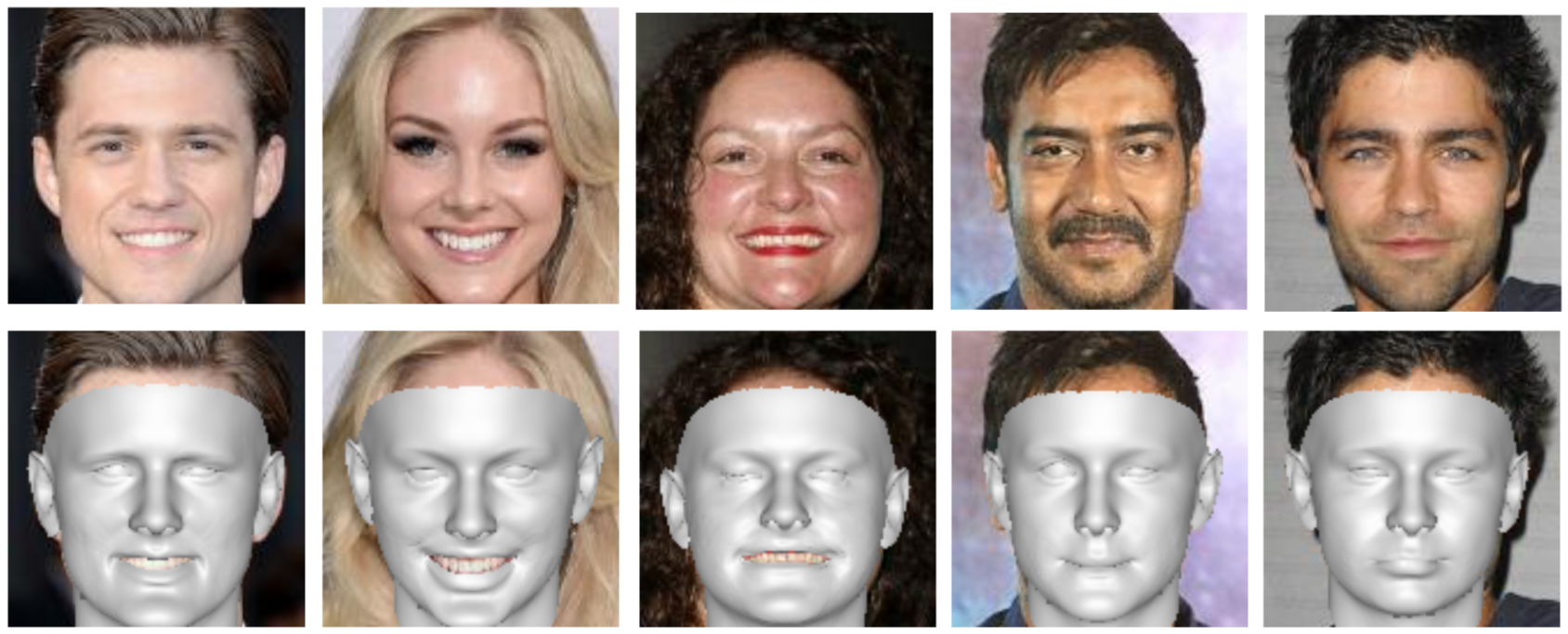}
\end{center}
\vspace{-15pt}
  \caption{\textbf{Samples of face meshes in Voxceleb-3D.} We overlay the 3D faces with associated images to show how well 3D meshes fit in 2D face outlines.}
  \vspace{-10pt}
\label{voxceleb-3d}
\end{figure} 

%Voxceleb-3D contains wider to thinner face shape variation.

\section{Experiments and Results}
\label{sec:exper}

\textbf{Datasets.} 
We use our created Voxceleb-3D dataset described in Sec. \ref{sec:methods/supervised}.
There are about 150K utterances and 140K frontal face images from 1225 subjects. 
The train/test split for Voxceleb-3D is the same as \cite{NEURIPS2019_eb9fc349}:
Names starting with A-E are used for testing, and the others are for training. 
We manually pick the best-fit 3D face models for each identity as reference models for evaluations. We display samples of face meshes in Fig. \ref{voxceleb-3d}.

\textbf{Data Processing and Training.} We follow \cite{NEURIPS2019_eb9fc349} and extract 64-dimensional log mel-spectrograms with a window size of 25 ms, and perform normalization by mean and variance of each frequency bin for each utterance. In the unsupervised setting, we adopt SynergyNet \cite{wu2021synergy} as the expert. Face images from the generator are 64$\times$64, and we bilinearly upsample them to 120 $\times$120 to fit the input size of the expert for 3D face reconstruction from images.
%\textbf{Network Architectures.} $\phi_v$ of both supervised and unsupervised settings and GAN are the same voice encoder  \cite{NEURIPS2019_eb9fc349} pretrained on the large-scale speech recognition task. For supervised learning, we adopt fully-connected layers to regress $\alpha$ from the voice embedding $v$. For unsupervised learning, GAN in the speech-to-image block follows the same structure of \cite{NEURIPS2019_eb9fc349}, and the image-to-3D-face block follows the expert \cite{wu2021synergy}. 
Our framework is implemented in PyTorch \cite{paszke2019pytorch}. We use Adam optimizer \cite{kingma2014adam} and set the learning rate to 2$\times$10$^{-4}$, batch size to 64, and a total number of training steps to 50,000, which consumes about 16 hours to train on a machine with a GeForce RTX 2080 GPU. 

To train with triplet loss, for each sample in a batch, we further uniformly sample one utterance of the same person as the positive sample and sample the other one of the different person as the negative sample. We illustrate the network architectures in the supplementary

%The groundtruth faces are cropped to 64$\times$64 

\textbf{Metrics.}
We design several metrics to evaluate 3D face deformation based on $\alpha$. Here we introduce a line-based metric, ARE, and present point-based and region-based metrics using iterative closet point registration and facial landmarks in the supplementary.

\begin{figure}[bt!]
\begin{center}
\includegraphics[width=0.35\linewidth]{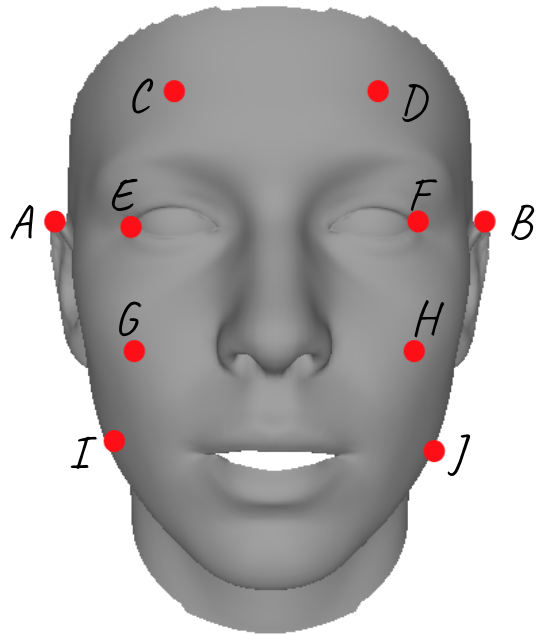}
\end{center}
  \vspace{-12pt}
  \caption{\textbf{Distance illustration for our ARE metric.} $\overline{AB}$: ear-to-ear distance. $\overline{CD}$: forehead width. $\overline{EF}$: outer-interocular distance. $\overline{GH}$: midline distance. $\overline{IJ}$: cheek-to-cheek distance.}
  \vspace{-10pt}
  \label{p2pDistance}
\end{figure}

Absolute Ratio Error (\textbf{ARE}, line-based): Distances between facial points are commonly used as measures related to aesthetics or surgical purposes \cite{sarver2007aesthetic, pallett2010new, abdullah2002inner}.  
We pick point pairs (shown in Fig. \ref{p2pDistance}) that are most representative for evaluation and calculate the distance ratios to outer-interocular distance (OICD). For example, ear ratio (ER) is $\overline{AB}/\overline{EF}$, and the same for forehead ratio (FR), midline ratio (MR), and cheek ratio (CR). 
We evaluate our models by the absolute ratio error (ARE) between the predicted and the reference face meshes
because these ratios can capture face deformation.
As an example, ARE of ER is $|\text{ER} - \text{ER}^*|$, where $^*$ denotes the ratios of reference models.

% 2. \textit{Point-to-Plane Root Mean Square Error} (\textbf{Point-to-plane RMSE}, region-based): We follow the surface registration for 3D models using the popular iterative closest point (ICP) \cite{pomerleau2015review} algorithm to align the predicted and reference meshes. We then calculate point-to-plane RMSE. Registration for both whole face and facial parts are considered in quantitative analysis (Sec.\ref{sec:num}).

\begin{figure}[bt!]
\begin{center}
\includegraphics[width=1.0\linewidth]{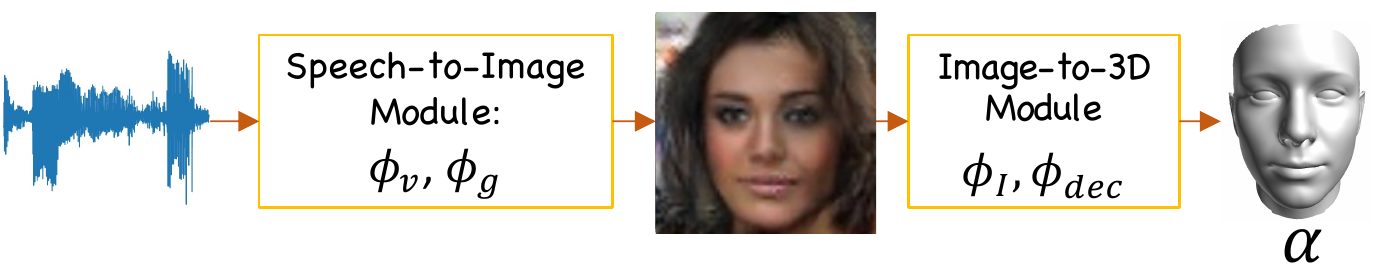}
\end{center}
  \vspace{-15pt}
  \caption{\textbf{Baseline framework.} The baseline is a direct cascade of two pre-trained state-of-the-art modules: speech-to-image \cite{NEURIPS2019_eb9fc349} and image-to-3D-face modeling \cite{guo2020towards,wu2021synergy}.}
  \setlength{\belowcaptionskip}{0pt}
  \vspace{-5pt}
\label{baseline}
\end{figure}

\textbf{Baseline}.
We build a straightforward baseline by directly cascading two separately pretrained methods without joint training: the GAN-based speech-to-image block \cite{NEURIPS2019_eb9fc349} and SynergyNet \cite{wu2021synergy} for image-to-3D-face block (illustrated in Fig. \ref{baseline}) to produce 3D meshes from voices as the baseline framework.
In addition, 3DDFA-V2  \cite{guo2020towards} is another method for 3D face modeling from monocular images using BFM and holds a close performance to SynergyNet. Thus, we experiment with combinations of speech-to-image block + 3DDFA-V2 (Base-1) and speech-to-image block + SynergyNet (Base-2).
Aside from network-based approaches, we also devise simple oracles that use mean shapes of labels, such as male/female, as predictions, and provide the results as references in the supplementary.

\begin{figure}[bt!]
\begin{center}
\includegraphics[width=1.0\linewidth]{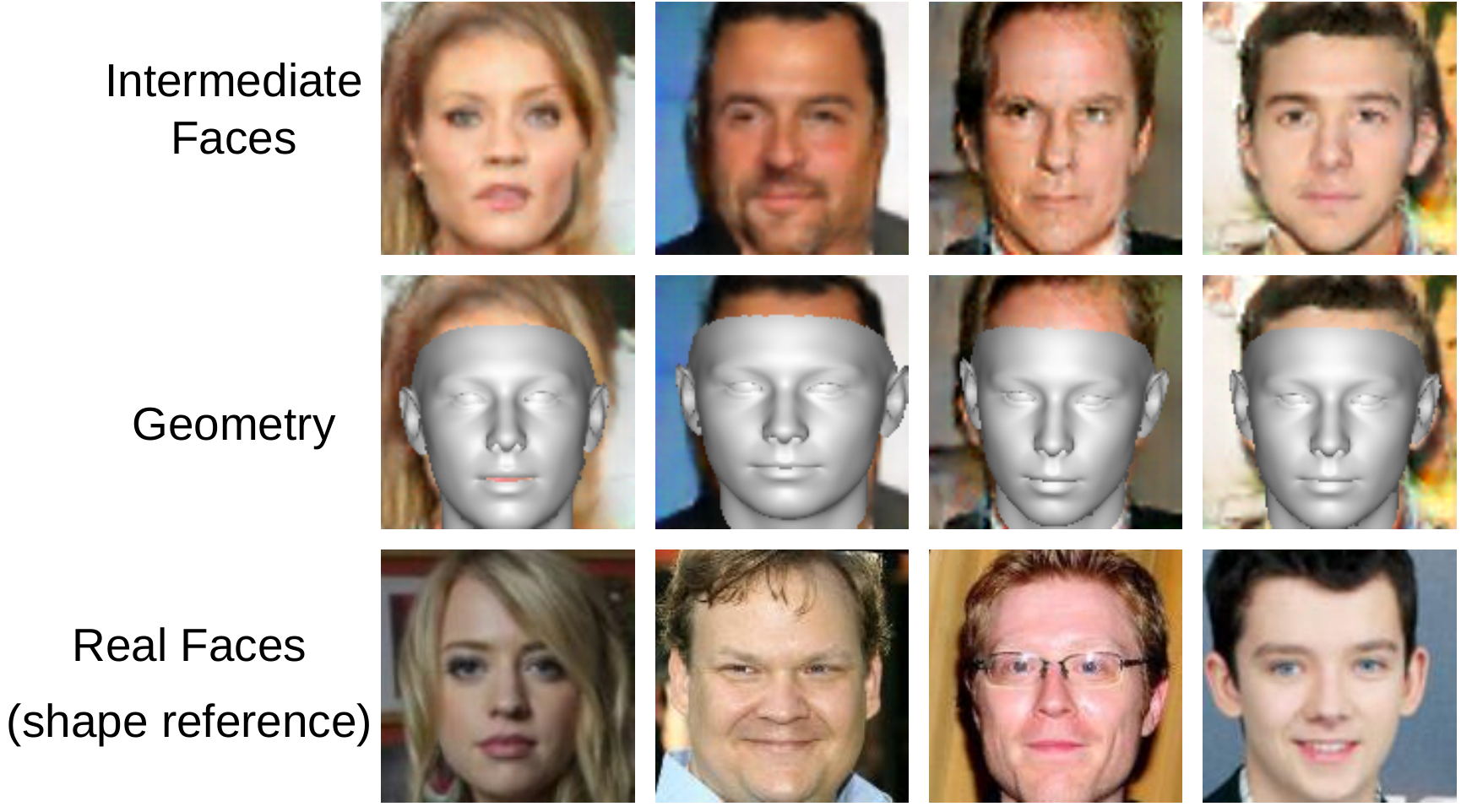}
\end{center}
  \vspace{-17pt}
  \caption{\textbf{Evidence for positive response to Q1}. Our unsupervised framework predicts intermediate 2D images and 3D meshes. This answers to Q1 that 3D face models exhibiting similar \textit{face shapes} to the references can be predicted from only voice inputs.}
  \vspace{-12pt}
\label{VGGFace_compare}
\end{figure} 

\begin{figure*}[hbt!]
\begin{center}
\includegraphics[width=1.0\linewidth]{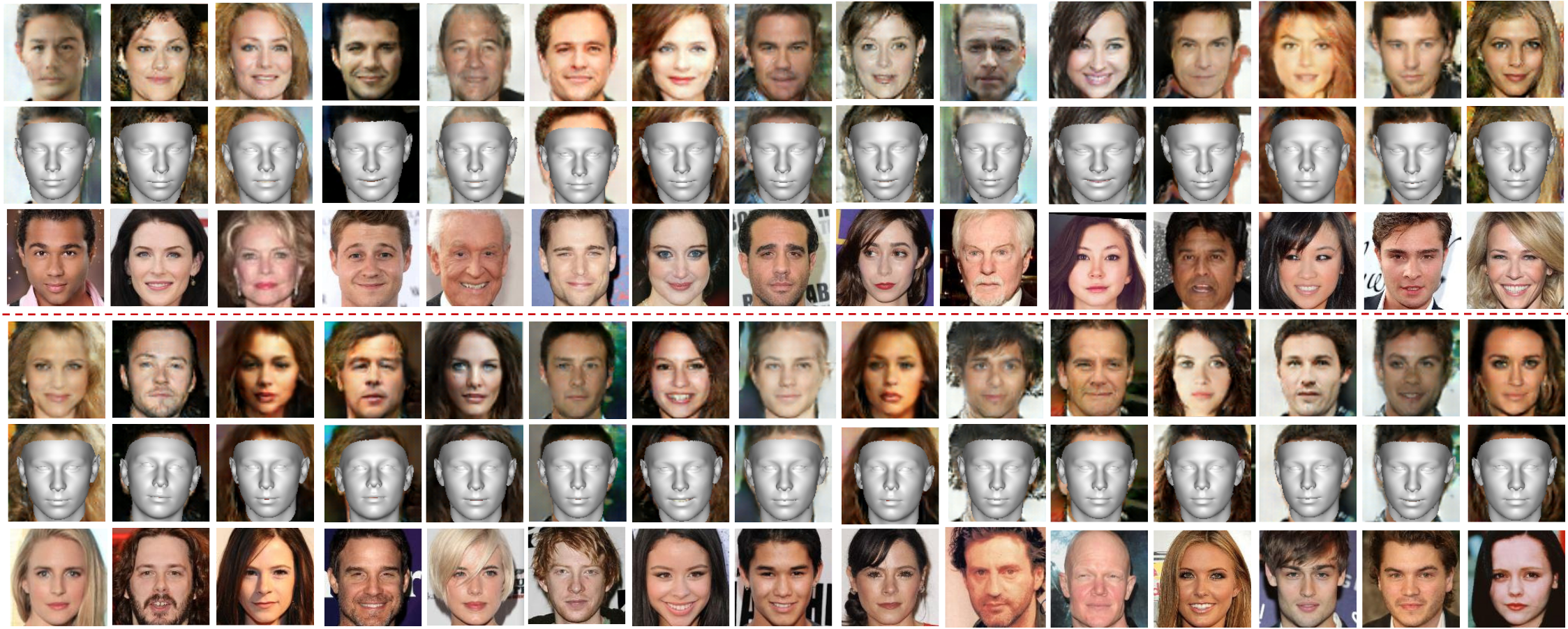}
\end{center}
  \vspace{-15pt}
  \caption{\textbf{A collection of results supports our positive response to Q1.} This figure extends Fig.\ref{VGGFace_compare}. Top to down for two row chunks: predicted intermediate face images, predicted 3D models, real faces for references.}
  \vspace{-15pt}
\label{additional}
\end{figure*}

\subsection{Analysis}
\label{sec:vis}
We attempt to answer Q1-Q4 raised in Sec.\ref{sec:intro} in this section and respond to each respective question in A1-A4. In A1-A3, we show predictions using our unsupervised learning setting since the by-product intermediate images help explain 3D mesh prediction for better comprehension of the mechanism. We show visuals from the supervised version in the supplementary.
 
\begin{figure}[bt!]
\begin{center}
\includegraphics[width=1.0\linewidth]{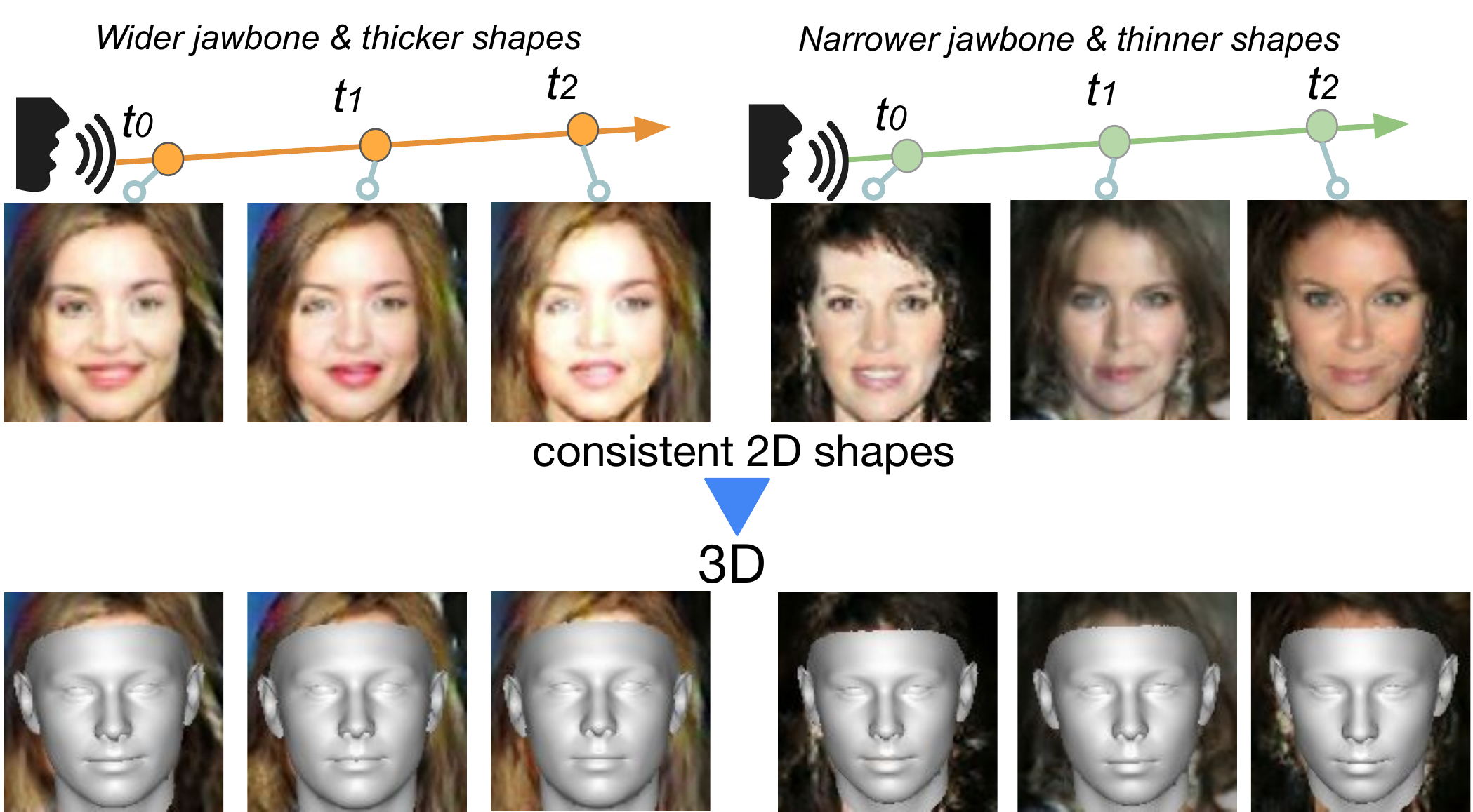}
\end{center}
    \vspace{-18pt}
    \caption{\textbf{Illustration for our positive response to Q2.} Consistent intermediate images and 3D faces can be predicted from the same speaker with different time-step utterances.}
    \vspace{-10pt}
\label{coherence}
\end{figure} 

\begin{figure}[bt!]
\begin{center}
\includegraphics[width=1.0\linewidth]{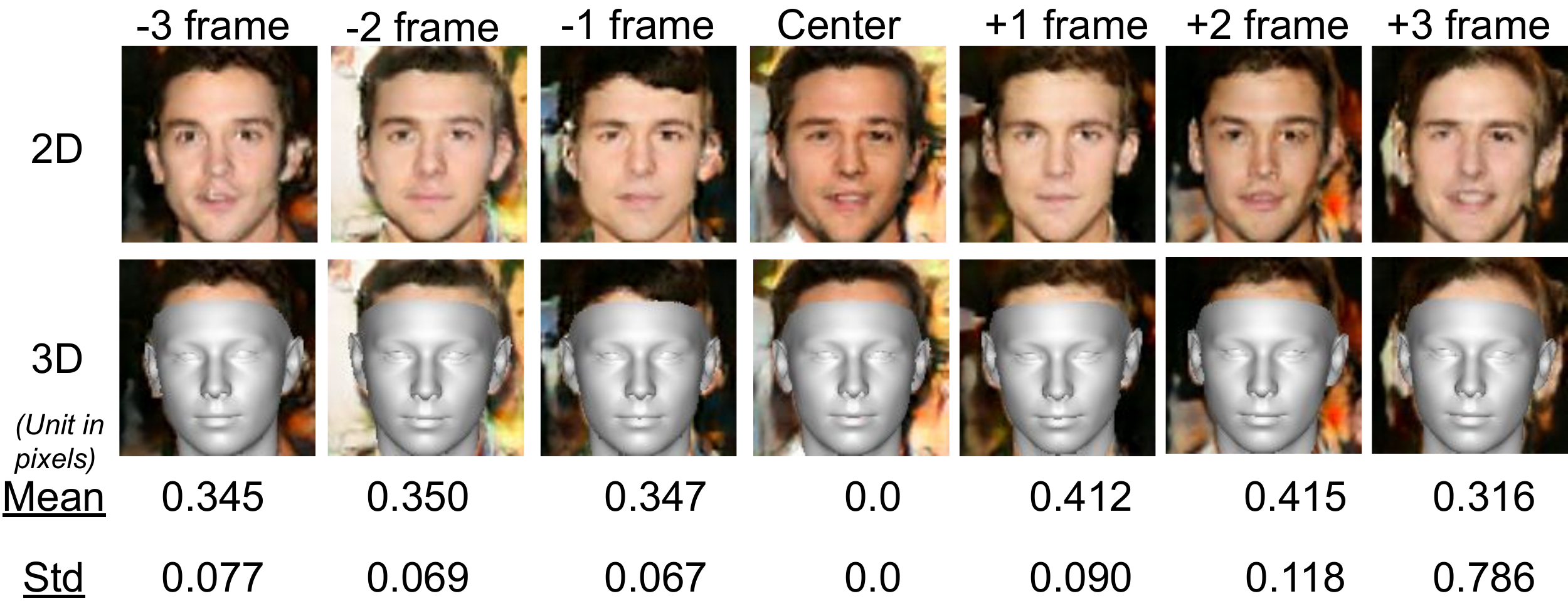}
\end{center}
    \vspace{-15pt}
    \caption{\textbf{Shape variation statistics in response to Q2.} Mean and std of per-vertex variation w.r.t. the center frame are shown, calculated in frontal pose. 3D shapes recovered from different utterances are consistent with only sub-pixel differences.}
    \vspace{-9pt}
\label{statistics}
\end{figure} 

\textbf{A1: Meshes and intermediate images.}
In Fig. \ref{VGGFace_compare}, we display intermediate 2D images, 3D meshes, and real faces. Note that the real faces should only be treated for identification purposes in terms of \textbf{face shapes} because those images include backgrounds or hairstyle variations that differ from references. Our end targets are the 3D face meshes that are free from these factors.
Prediction from our framework generates wider meshes in Column 2 and thinner meshes in Column 3 and 4, which reflect the real face wideness.
All the generated 3D meshes fit in 2D facial outlines well.

These results exemplify the ability to convert voices into plausible 3D face meshes. Although meshes are rough compared with 3D synthesis from images or videos modalities, the results conform to our intuitions that when an unheard speech comes, one can roughly envision whether the speaker's face is overall wider or thinner. However, we cannot picture subtle details, such as bumps or wrinkles on faces. The same trends can be observed in a vast result collection in Fig. \ref{additional}. The results are not cherry-picked.

\textbf{A2: Prediction coherence of the same speaker.}
To address Q2, we showcase in Fig. \ref{coherence} and \ref{statistics} for the coherence of the predicted face shapes from different utterances of the same speakers.
The 2D predictions exhibit \textit{face shape and outline consistency},
though they are still plagued by stylistic variations that are geometrically unrelated to our task.
This not only confirms the ability to produce coherent face meshes but also underlines why predicting face meshes from voices is regarded as less noisy than face image synthesis.

\begin{figure}[bt!]
\begin{center}
\includegraphics[width=1.0\linewidth]{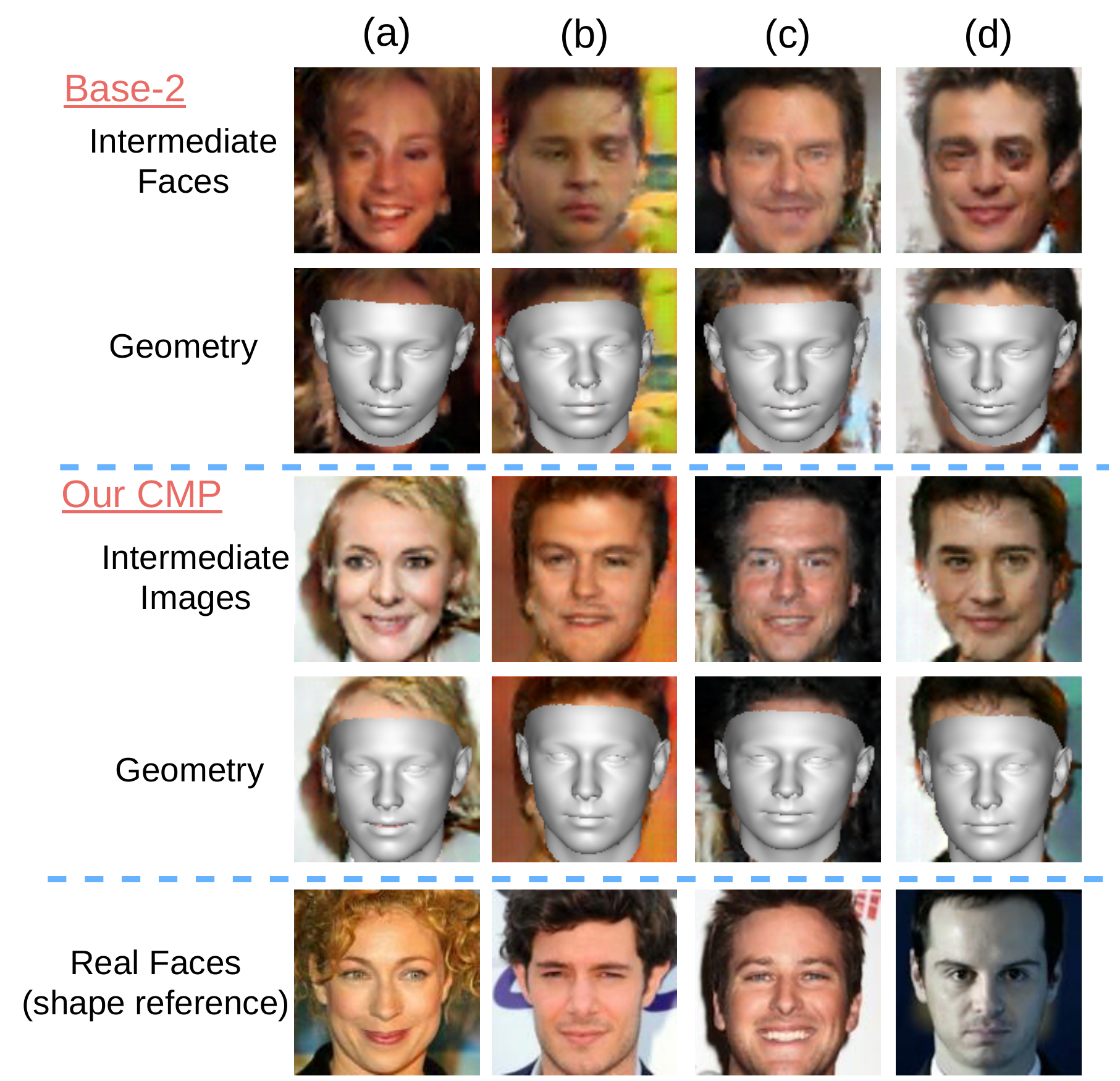}
\end{center}
  \vspace{-15pt}
  \caption{\textbf{Comparison of intermediate images and meshes in response to Q3.} The cross-modal joint training strategy in our unsupervised CMP produces better-quality images than the baseline. More reliable images as latent representations from our CMP can facilitate the mesh prediction. We include real faces for \textit{face shape} references.}
  \vspace{-10pt}
\label{baseline_comp}
\end{figure}

\textbf{A3: Gain from cross-modal joint training.}
For Q3, we compare results from our unsupervised framework against those from the baseline in Fig. \ref{baseline_comp}.
Joint training for the speech-to-image and image-to-3D sub-networks attain higher and more stable image synthesis quality, which benefits 3D mesh prediction. In contrast, those from the baseline (Base-2) include more artifacts. This justifies our CMP's cross-modal joint training strategy, which lets networks learn to predict 3D faces with voice input at the training, improves over the baseline that is separately trained.

To this end, we understand that voices can help 3D face prediction and produce visually reasonable meshes that are close to real face shapes.

\textbf{A3-Quantification+A4}.
We numerically compare supervised and unsupervised settings of our analysis framework, CMP, against the baseline (Fig. \ref{baseline}) using the ARE proposed in Sec.\ref{sec:exper}. 
Both supervised and unsupervised settings improve the line-based ARE over the baseline around \textit{20\%}, as exhibited in Table \ref{ARE_metric}.
The results show that cross-modal joint training achieves better results than the direct cascade of pretrained blocks.
These improvements reveal underlying correlations between voices and face shapes such that training face mesh prediction with joined voice information is helpful.
Among all metrics, ear ratio (ER) has the most prominent improvements, 
indicating that the best indicative attribute voice can hint is the head width, and thus it answers Q4.
This analysis aligns with the findings in Sec.\ref{sec:vis} that voice can indicate wider/thinner faces, which corresponds to our intuition that we can roughly envision a speaker's face width from voices.
Through this study, we quantify the improvements of cross-modal learning from voice inputs, and the findings echo human perception intuitively.

\begin{table}[tb!]
\begin{center}
 \setlength{\abovecaptionskip}{3pt} 
  \caption{\textbf{ARE metric study.} Compared with baseline in Fig. \ref{baseline}, results from CMP show that cross-modal joint training with voice input can obtain around 20\% improvements. We also highlight the largest improvement, ER, that answers to Q4.}
  \small
  \label{ARE_metric}
  \begin{tabular}[c]
  {|
  p{1.10cm}<{\centering\arraybackslash}|
  p{1.0cm}<{\centering\arraybackslash}|
  p{1.0cm}<{\centering\arraybackslash}|
  p{1.5cm}<{\centering\arraybackslash}|
  p{1.5cm}<{\centering\arraybackslash}|}
  \hline
      ARE   & Base-1 & Base-2 & CMP-supervised & CMP-unsupervised  \\
    \hline
       ER & 0.0319 & 0.0311 & \cellcolor{yellow!30}\textbf{0.0152} & \cellcolor{yellow!30}\textbf{0.0181} \\ %[3.813, 4.420, 2.888, 3.707]
       FR & 0.0184 & 0.0173 & 0.0186 & 0.0169\\ 
       % [3.718, 4.394, 2.852, 3.655] 
       MR & 0.0177 & 0.0173 & 0.0169 & 0.0174 \\ 
       %  [3.718, 4.368, 2.877, 3.654] 
       CR & 0.0562 & 0.0551 & 0.0457 & 0.0480 \\
       \hline
       Mean & 0.0311 & 0.0302 & \textbf{0.0241} & \textbf{0.0251}  \\
       Gain & - & 0\% & \cellcolor{blue!25}\textbf{-20.2}\% & \cellcolor{blue!25}\textbf{-16.9}\% \\
    \hline
  \end{tabular}
  \vspace{-11pt}
\end{center}
\end{table}

\newcommand {\Hnull} {\mathcal{H}_0}
\newcommand {\Halt} {\mathcal{H}_1}

\begin{figure}[t!]
\begin{center}
\includegraphics[width=0.92\linewidth]{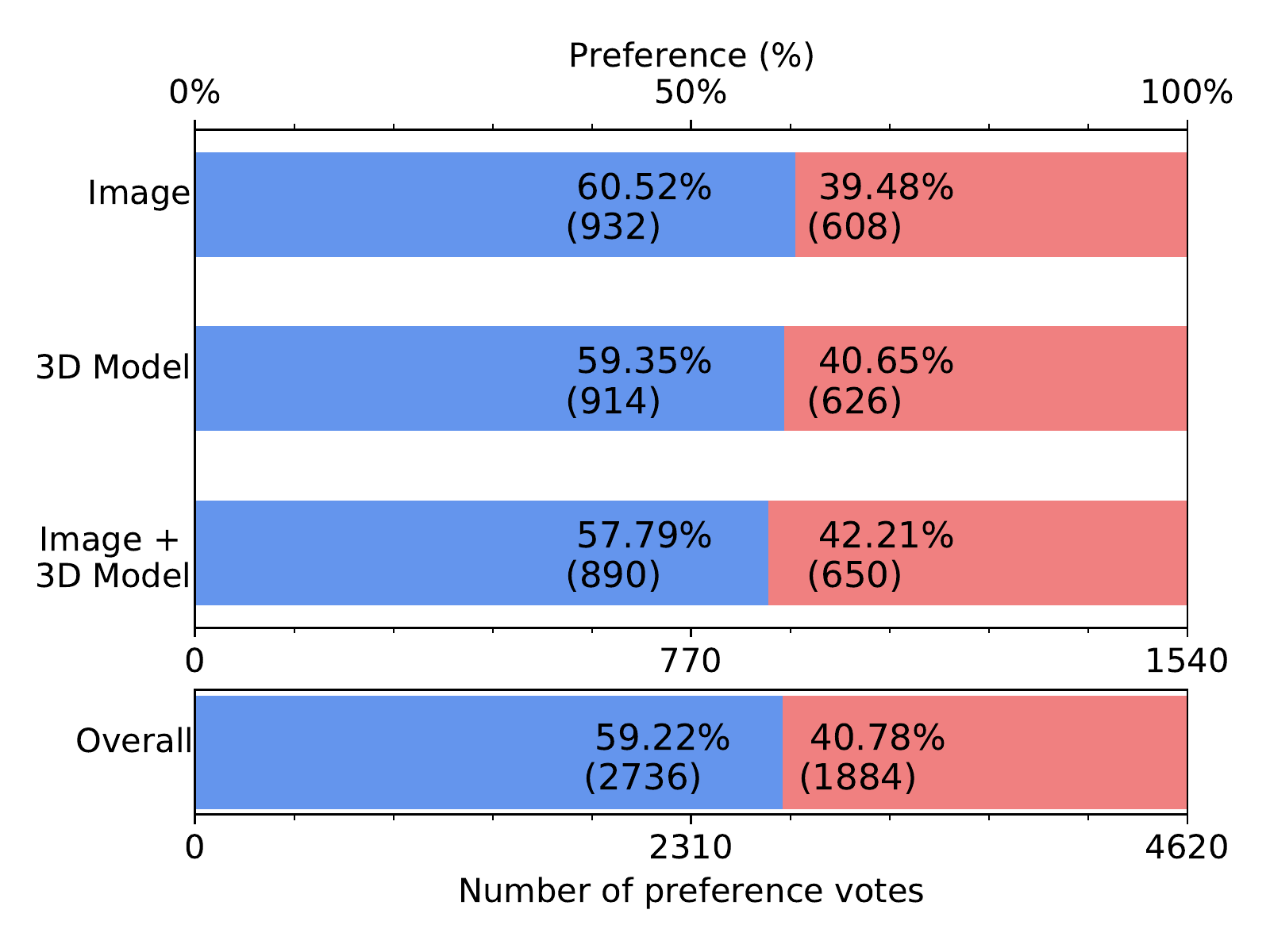}
\small
\begin{tabular}[c]
  {|
  p{1.0cm}<{\centering\arraybackslash}|
  p{0.95cm}<{\centering\arraybackslash}|
  p{1.5cm}<{\centering\arraybackslash}|
  p{1.5cm}<{\centering\arraybackslash}|
  p{1.1cm}<{\centering\arraybackslash}|}
  \hline
         & Image & 3D Model & Image + 3D Model & Overall  \\
    \hline
       $p$-value& \begin{small}$\sim$10$^{-16}$ \end{small} & \begin{small}$\sim$10$^{-14}$ \end{small} & \begin{small}$\sim$10$^{-10}$ \end{small} & \begin{small}$\sim$10$^{-16}$ \end{small} \\
    \hline
  \end{tabular}
\end{center}
  \vspace{-12pt}
  \caption{\textbf{Results of subjective preference tests.} 
  The blue bars are the preference for our method, while the red bars are the preference for the baseline method.
  The percentages are labeled on the bar, and the total number of votes is enclosed in the parentheses.
  The $x$-axis on the bottom labels the total number of responses, and that on the top denotes the percentage. The $p$-values of the statistical significance tests are provided under the bar. $\sim$ shows the value's order of magnitude.}
\label{fig:pref-test}
\vspace{-10pt}
\end{figure}

\subsection{Subjective Evaluations}
\label{sec:human_judgement}
We further conduct subjective preference tests over the outputs to quantify the difference of preference.
The test was divided into three sections, considering \textit{images}, \textit{3D models}, and \textit{joint materials}. Though we favor face meshes over images because the former are free from irrelevant textures or backgrounds, we included intermediate images from our unsupervised setting in the test and asked subjects to focus on face shapes since better-outlined shapes on images lead to better-shaped meshes, as indicated in Fig. \ref{baseline_comp}.

% (as mentioned in our paper Fig. 10 that more reliable latent images can facilitate 3D mesh prediction). Therefore we also conducted preference tests for face shapes on images. 

\textbf{Evaluation design.} Thirty questions were included in the test, and 154 subjects with no prior knowledge of our work were invited to the test.
In the first section, each of the ten questions consisted of three images-- a reference face image, a face image from our unsupervised CMP, and a face image generated from the Base-2 (\cite{NEURIPS2019_eb9fc349}+\cite{wu2021synergy}).
The order of the generated images was randomized.
The subjects were asked to select the face image "whose shape is geometrically more similar to the reference face?".
In the second section (10 questions), a similar design was laid out, but 3D face models from Base-2 and our CMP were used instead of images.
Finally, in the third section (10 questions), each of the two options comprised a face image and a 3D face model; the subject was asked to jointly consider over the two materials: "overall, whose shape geometrically fits the given reference image better?"

\textbf{Statistical significance test.} Fig. \ref{fig:pref-test} summarizes our subjective evaluation.
We conduct a statistical significance test with the following formulation.
A subject's response to a question is considered as a Bernoulli random variable with a parameter $p$.
The null hypothesis ($\Hnull$) assumes $p \le 0.5$, meaning that the subjects do not prefer our model.
The alternative hypothesis $\Halt$ assumes $p > 0.5$, meaning that the subjects prefer our model.
For each section, there are 154 subjects and ten responses per subject.
For a significance level $\gamma=0.001$, let $b_{n,p}(\gamma)$ denote the quantile of order $\gamma$ for the binomial distribution with parameters $p$ and $n$.
We can decide whether the subjects prefer our model by
\begin{equation}
\begin{split}
    &\mathrm{Reject ~\Hnull ~versus~ \Halt \Leftrightarrow} ~np \ge b_{n,p}(1 - \gamma) \\
    & \Hnull: p \le 0.5, \Halt: p > 0.5.
\end{split}
\end{equation}
As shown in Fig. \ref{fig:pref-test}, $np$ is well above the threshold $b_{n=1540,p=0.5}(1 - \gamma) = 831$, rejecting $\Hnull$ and suggesting that the subjects significantly prefer our model over the baseline.
The single-sided $p$-values are displayed under the bar chart. A lower $p$-value means stronger rejection of $\Hnull$. The $p$-values from our tests are much lower than the level 0.001, showing high statistical significance.
In conclusion, the hypothesis test verifies that the subjects indeed favor the predictions from our method.
\section{Conclusion and Discussion}
\label{sec:conclusion}
In this work, we investigate a root question in human perception: can face geometry be gleaned from voices? We first point out shortcomings in previous studies in which 2D faces are predicted: such synthesis contains variations in hairstyles, backgrounds, and facial textures with controversial correlations to voices. We instead focus on 3D faces whose correlations to voices have been supported by neuroscience and cognitive science studies. As a pioneering work toward this direction, we innovate a way to construct Voxceleb-3D that includes paired voices and 3D face models, devise and test baseline methods and oracles, and propose a set of evaluation metrics. Our proposed main framework, CMP, learns 3DMM parameters from voices under both supervised and unsupervised settings. Based on CMP, we answer the core question with a four-part breakdown by detailed analyses and subjective evaluations. We conclude that 3D faces can be roughly reconstructed from voices.
Our study is far from complete, but hopefully, it lays a foundation for speech and 3D cross-modal studies in the future.

\textbf{Ethical statement.}
There are arguably implicit factors, such as voices after smoking and drinking might be different. The data of Voxceleb contains speech from interviews, where interviewees usually speak in normal voices. More implicit and subtle factors such as drug use or health conditions might affect voices, but it needs clinical studies and should be validated from physiological views.
The results shown in this work only aim to point out the correlation between voice and face (skull) structure exist and do not make assumptions on race/ethnic origin, and this work does not indicate the relation between race and voice or race and face structure. As mentioned in Introduction, the correlation between race/ethnicity cannot be easily resolved. Besides, the reconstructed meshes do not contain skin color, facial textures, or hairstyles that can explicitly correspond to one’s true identity, and thus anonymity can be preserved.

%Cross-modal learning is a focus of the machine learning and AI community. This work goes beyond previous 2D face image synthesis that may include irrelevant or controversial variations, such as hairstyles, backgrounds, and skin colors. We provide analyses focusing on the correlations between voices and face shapes. Our work suggests such correlations exist, and we contribute more tangible and explainable grounds of cross-modal learning to the community of interest.

%%%%%%%%% REFERENCES
{
    % \clearpage
    \small
    \bibliographystyle{ieee_fullname}
    \bibliography{macros,main}
}

% --- supplementary material
% \input{sec/X_supplementary}

% --- uncomment this to read the instructions
% \input{sec/X_instructions}

\clearpage
\newpage
\newpage
\pagebreak

\setcounter{section}{0}
\setcounter{equation}{0}
\setcounter{figure}{0}
\setcounter{table}{0}

\appendix

\renewcommand\thesection{\Alph{section}}
\renewcommand\thesubsection{\thesection.\Alph{subsection}}

\renewcommand{\theequation}{S-\arabic{equation}}
\renewcommand{\thefigure}{S-\arabic{figure}}
\renewcommand{\thetable}{S-\arabic{table}}

% --- PDF will be split by an editor (e.g. macOS preview), so need to restart from page 1
\setcounter{page}{1}

% --- repeat the title (AT: haven't found a more elegant way to do this...)
\twocolumn[
\centering
\Large
\textbf{Supplementary Material} \\
\vspace{1.0em}
] %< twocolumn
\appendix

\begin{figure}[hbt]
\begin{center}
\includegraphics[width=1.0\linewidth]{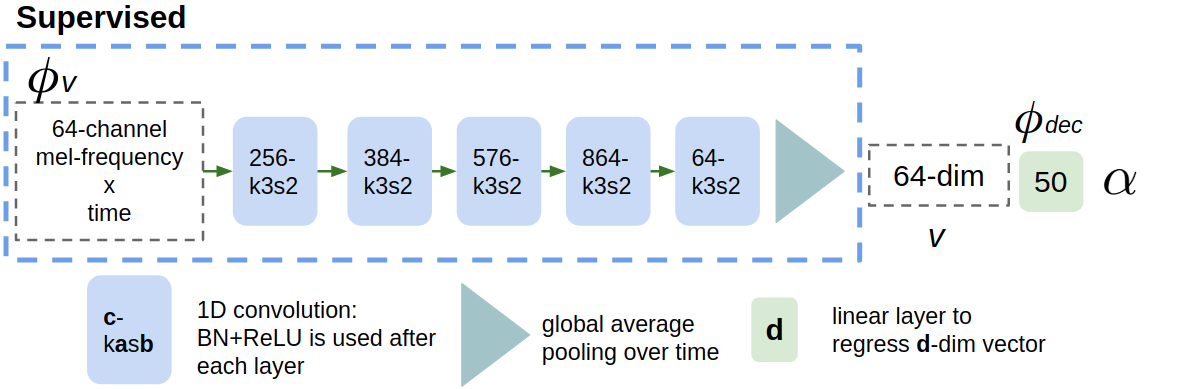}
\vspace{3pt}
\includegraphics[width=1.0\linewidth]{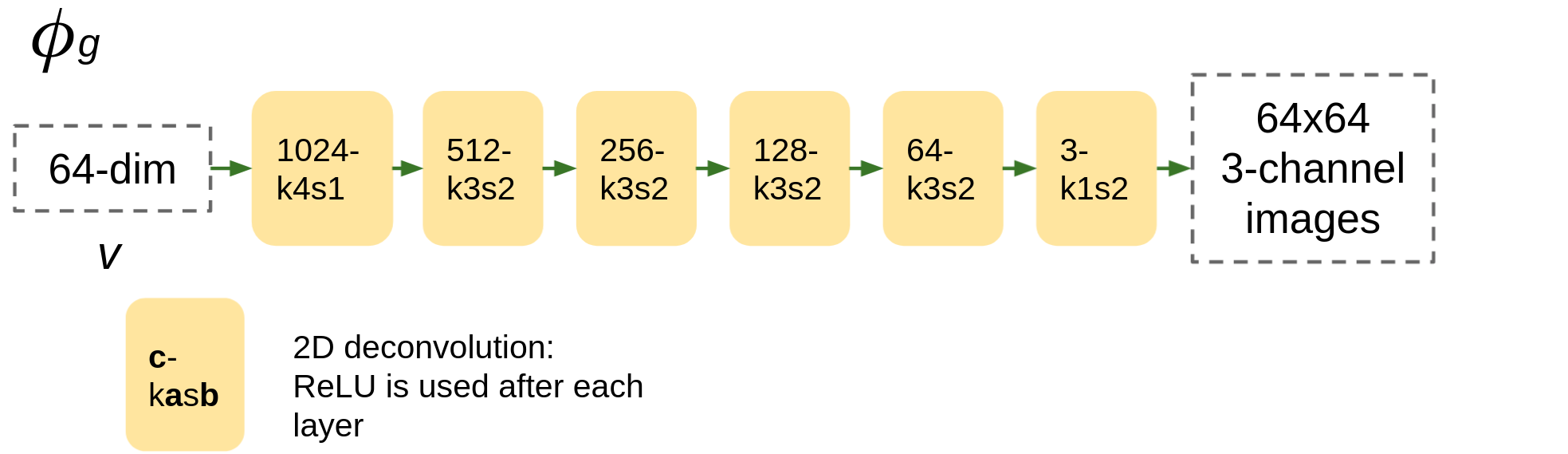}
\vspace{3pt}
\includegraphics[width=1.0\linewidth]{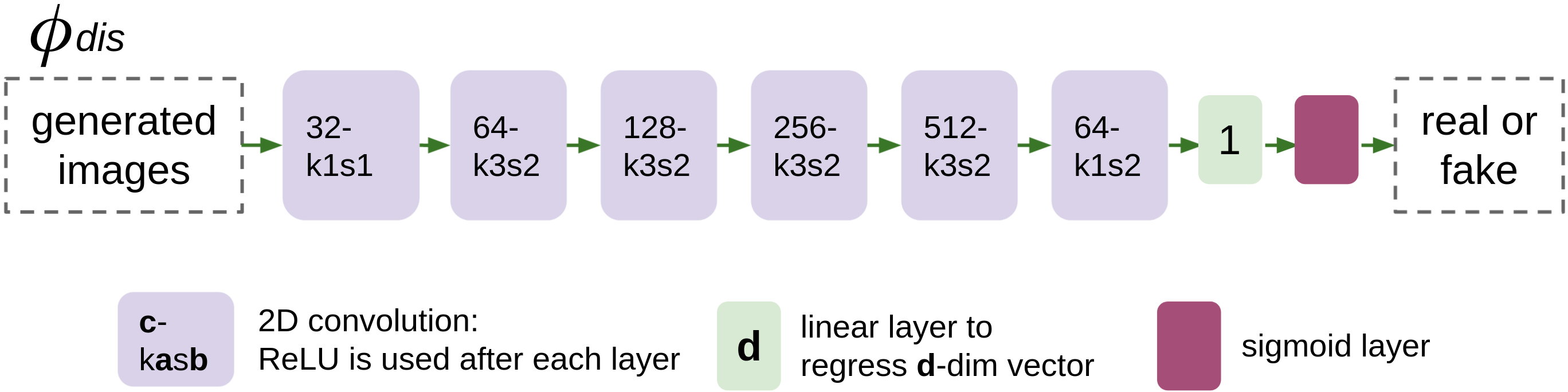}
\vspace{3pt}
\includegraphics[width=1.0\linewidth]{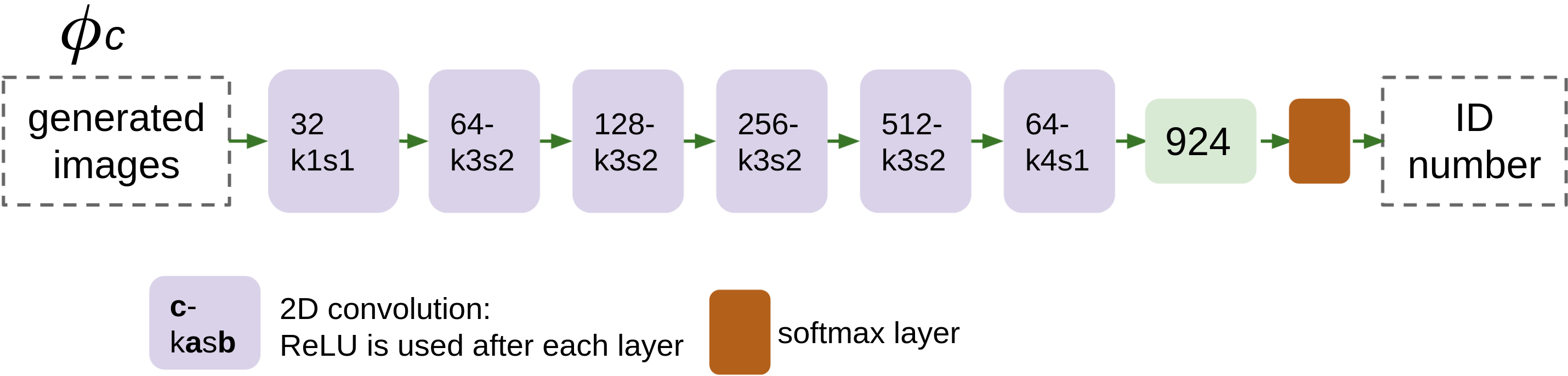}
\vspace{3pt}
\includegraphics[width=1.0\linewidth]{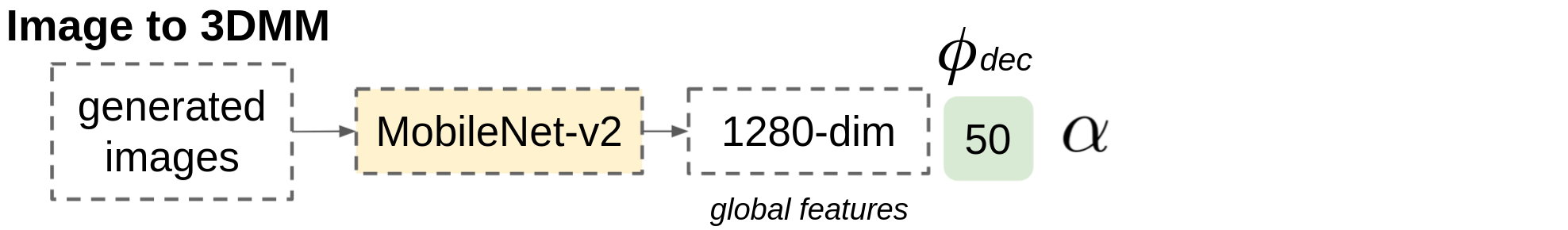}
\end{center}
  \vspace{-6pt}
  \caption{\textbf{Illustrations of network architectures.} \textbf{c}-k\textbf{a}s\textbf{b} means a convolutional layer with a \textbf{c}-channel output using a kernel size \textbf{a} and stride \textbf{b}. \textbf{d} in the linear layer means to output a \textbf{d}-dimension vector. The first graphic contains our supervised learning framework with $\phi_v$ and $\phi_{dec}$. The unsupervised setting contains $\phi_g$, $\phi_{dis}$, $\phi_c$, $\phi_v$ (the same structure as in the supervised setting), and the image-to-3DMM module. }
  
\label{network_arch}
%\vspace{-7pt}
\end{figure} 

\section{Overview}

This supplementary document is organized as follows. 
In Sec. \ref{sec:net_arch}, we show detailed network architectures for both supervised and unsupervised learning scenarios. 
In Sec. \ref{sec:dataset}, we describe details about our Voxceleb-3D training and evaluation split.
In Sec. \ref{sec:vis_supervised}, we respond to Q1-Q4 using illustrations from our CMP- supervised learning setting. 
In Sec. \ref{sec:point-based}, we introduce both point-based and region-based metrics and compare results produced from our CMP with those from baselines.
In Sec. \ref{sec:oracles}, we provide two simple oracles as non-network-based solutions for references using averaged face meshes in the training data as the predictions.
In Sec. \ref{sec:pose}, we show the robustness of head pose estimation of the expert network.
In Sec. \ref{sec:study_unsupervised}, we present more results from our unsupervised setting.
%In Sec. \ref{sec:facialPoints}, we show regions of facial parts for numerical evaluation in our paper Table 3.
In Sec. \ref{sec:applications}, we describe more on the applications of the cross-modal learning from voices to 3D faces. 
In Sec. \ref{sec:limitation}, we describe limitations of this work.
% Last, we describe our computing infrastructure in Sec. \ref{sec:infra}.

The numbering of figures, tables, and equations in this supplementary document go with the prefix 'S-'.

\vspace{-3pt}

\section{Network Architecture}
\label{sec:net_arch}
Here we exhibit detailed network architectures for both supervised and unsupervised settings in our CMP in Fig. \ref{network_arch}.

%\clearpage

\section{Voxceleb-3D Training/Evaluation Split}
\label{sec:dataset}
We display the details of the training/evaluation split in Table \ref{statistics}. As described in our paper Sec. 3.2 and Sec. 4-Datasets, Voxceleb-3D inherited from Voxceleb and VGGFace contains 1225 people. Names starting with 'A' - 'E' are included in the evaluation set, and the others are in the training set. 
The training set contains as many utterances we can fetch from Voxceleb, and the evaluation set contains three utterances for each person, amounting to a total of 0.9K utterances.
Face images are not included in the evaluation set because they cannot be used to calculate 3D face modeling errors, and thus we put a '-' mark in the table.

For 3D faces, we fit landmarks from images and obtain the optimized 3DMM and reconstructed 3D faces, as described in our paper Sec. 3.2 and Sec. 4-Datasets. There are several images associated with a person in VGGFace. We fit 3DMM parameters and reconstruct 3D meshes from these images. To create \textit{reference face meshes} for a person to fulfill quantitative evaluation, we manually select one neutral 3D face from the pool that best fits 2D facial outlines on images. Therefore, there are 301 3D face meshes represented in 3DMM parameters for each person as the reference. 

At test time, three utterances for each identity are used as inputs to reconstruct 3D faces. Those three predicted models are then used to compute quantitative results with the picked one reference model for each identity. 

Note that Voxceleb collects speech clips of interviews or talks for celebrities scraped from the web, and only gender labels are available in Voxceleb. Other features may require self-disclosure or are hard to trace, such as ages at the time of speaking, and thus are unavailable.

\begin{table}[htb]
\begin{center}
  \caption{\textbf{Voxceleb-3D training/evaluation split.} We provide data split details including number of utterances, number of face images, number of 3DMM parameters (equivalent to the number of 3D meshes), number of male and female, and number of people. Images for the evaluation set are not used for quantitative evaluation, and thus we mark the number '-'. We also display gender pie charts below the table.}
  \label{statistics}
  \vspace{-10pt}
  %\footnotesize
  \begin{tabular}[c]
  {|
  p{4.5cm}<{\centering\arraybackslash}|
  p{1.5cm}<{\centering\arraybackslash}|
  p{1.4cm}<{\centering\arraybackslash}|}
  \hline
       Dataset  & Training & Evaluation  \\
    \hline
       \# of utterances & 113K & 0.9K  \\
       \# of face images & 107K & -  \\
       \# of 3DMM param (face mesh) & 107K & 301  \\
       \# of male/female & 485/439 & 182/119 \\
       \# of people & 924 & 301 \\
    \hline
  \end{tabular}
  \vspace{-20pt}
  \begin{subfigure}{0.45\linewidth}
  \centering
  \includegraphics[width=\textwidth]{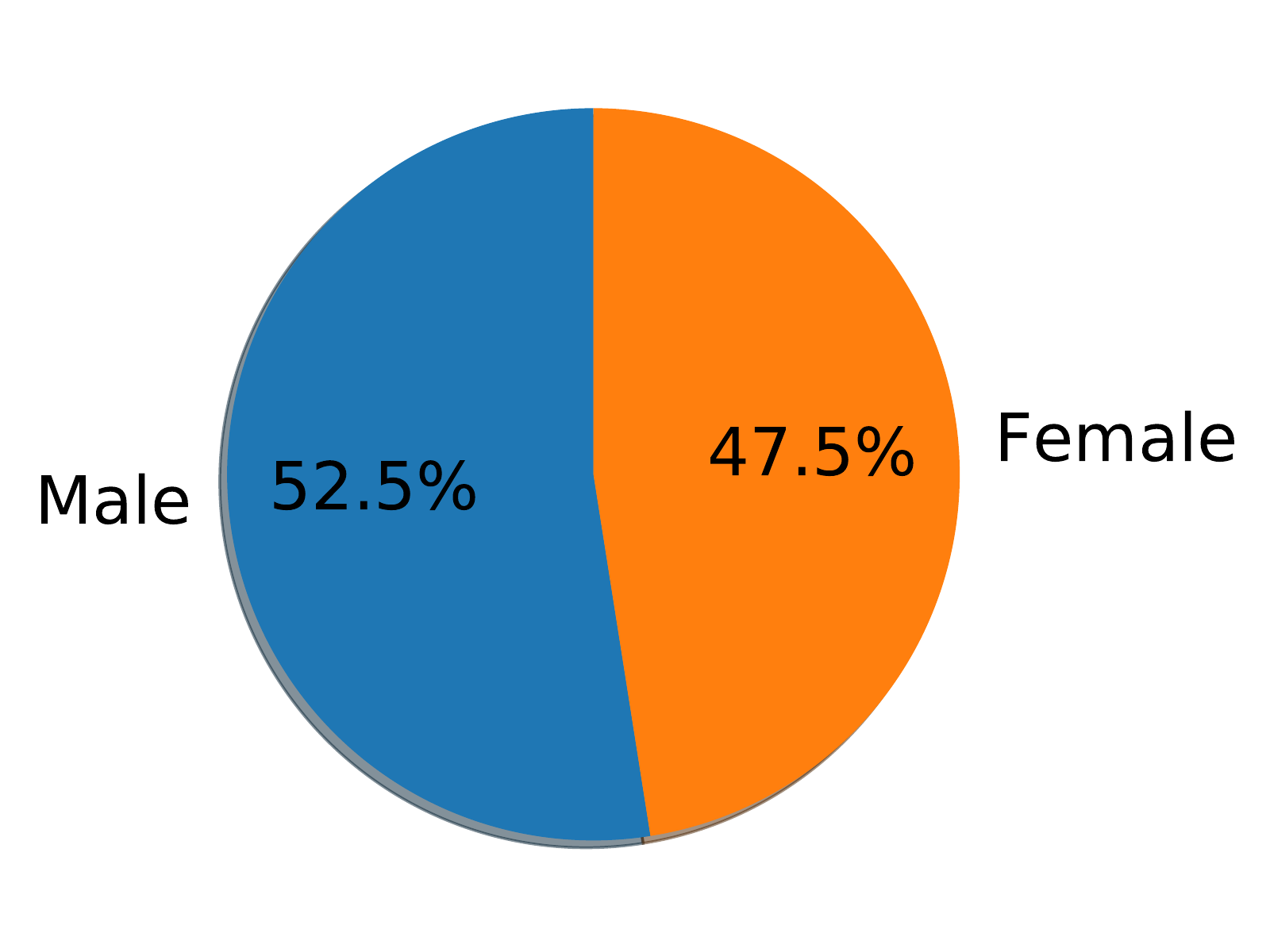}
  \vspace{-20pt}
  \caption{Training split}
  \label{fig:sub1}
\end{subfigure}%
\begin{subfigure}{0.45\linewidth}
  \centering
  \includegraphics[width=\textwidth]{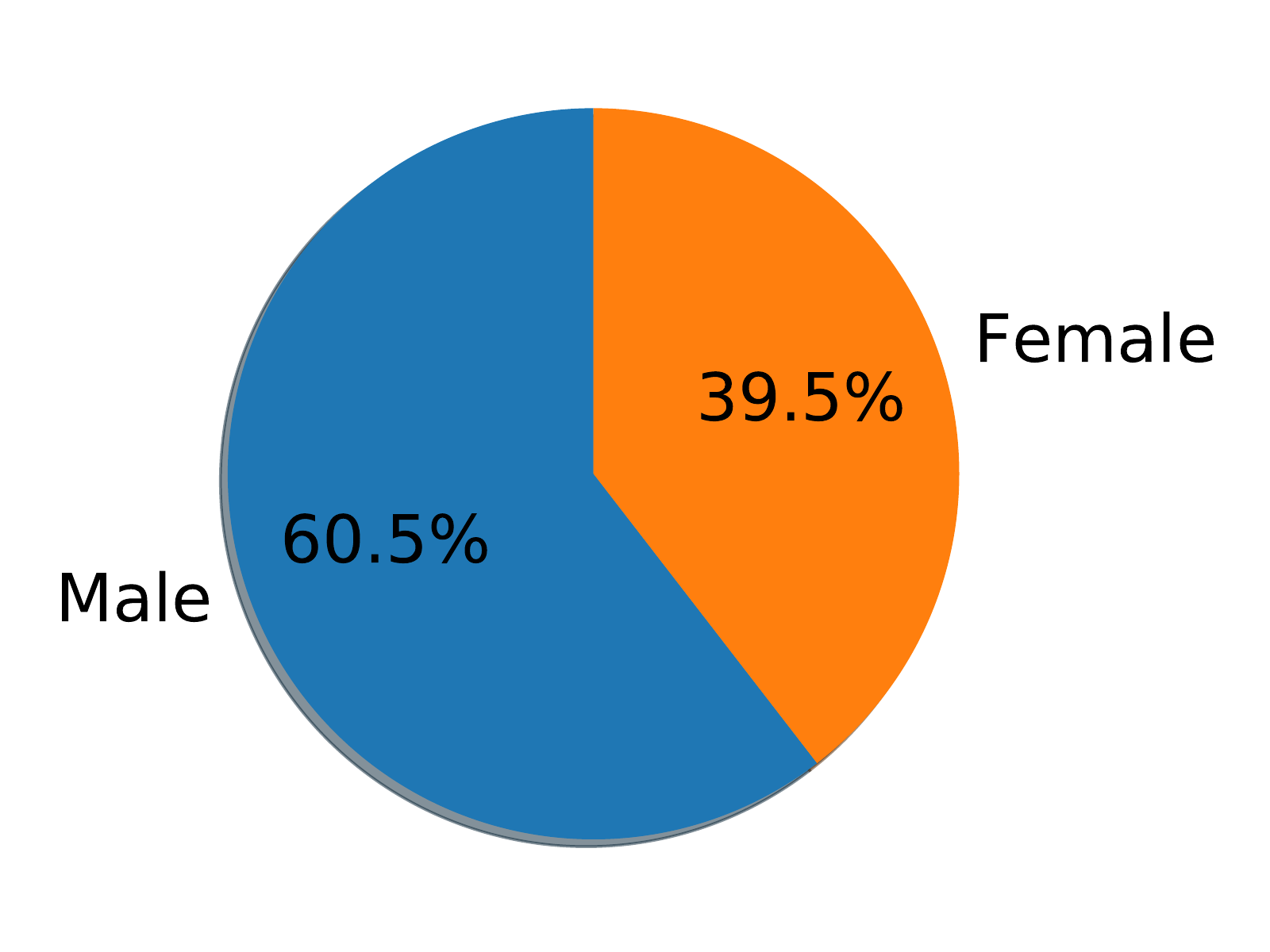}
  \vspace{-20pt}
  \caption{Evaluation split}
  \label{fig:sub2}
\end{subfigure}
\end{center}
\end{table}

\section{Response to Q1-Q4 using Supervised CMP}
\label{sec:vis_supervised}
Following main paper Sec.4.1-Analysis, we present the counterpart of A1-A4 using our supervised framework.

\begin{figure}[b]
\begin{center}
\includegraphics[width=1.0\linewidth]{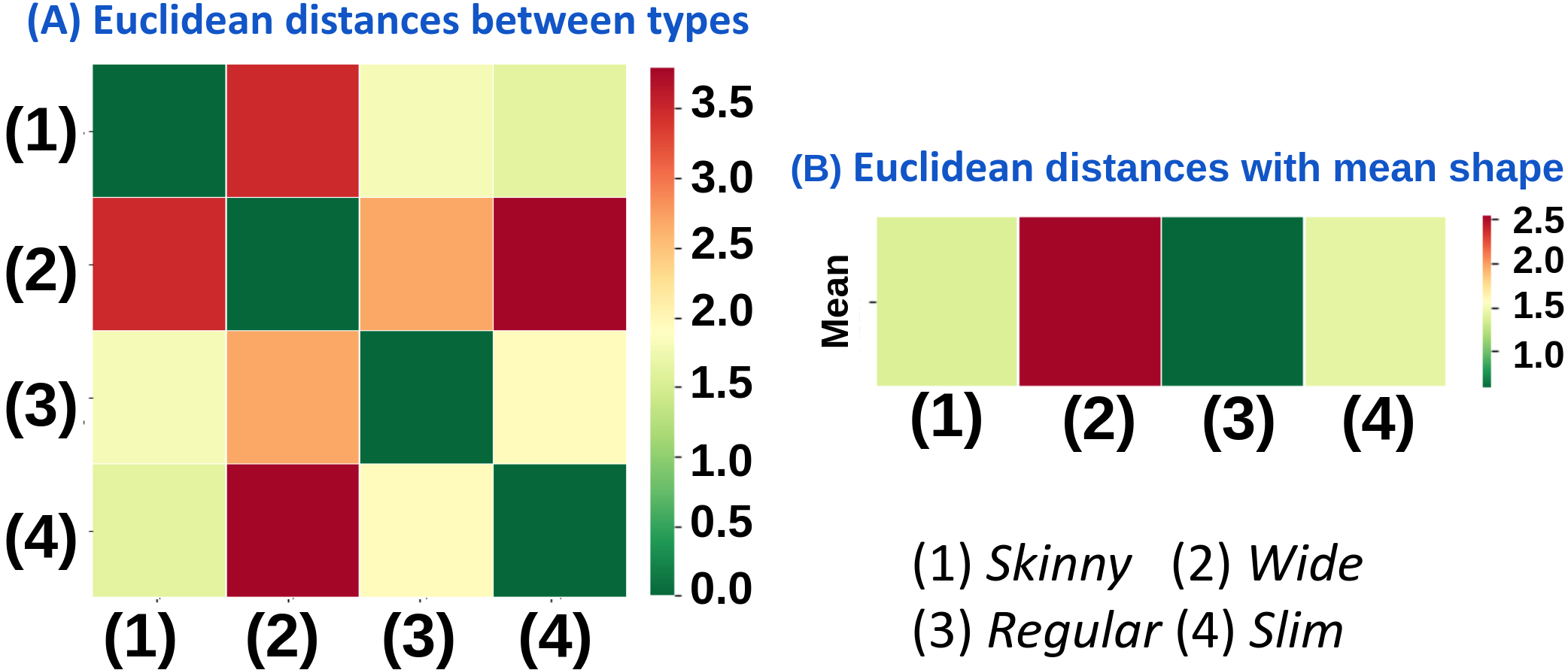}
\end{center}
  \vspace{-8pt}
  \caption{\textbf{Proximity for four face types.} We show (A) a confusion matrix and (B) distances with mean face shape in 3DMM parameter space to help comprehension for the face type variation  in Fig. \ref{supervised_gt}.}
\label{confusion}
\vspace{-6pt}
\end{figure}

\textbf{A1-Face meshes from our supervised learning.} In Fig. \ref{supervised_gt}, we present four types of face shapes -- \textit{skinny}, \textit{wide}, \textit{regular}, and \textit{slim} -- and show the reference images.
The produced face meshes from our supervised learning setting exhibit the model's ability to produce various types of face shapes and are also consistent with the reference images, which are provided for shape identification purposes. 
This illustration also validates our findings in Table 1 of the paper: the lowest absolute ratio error is ear-to-ear ratio (ER) distance, which is associated with overall face shapes, indicating wider or thinner faces.
We further investigate the proximity of the illustrated four face types, we calculate the parameter space Euclidean distances and show the confusion matrix in Fig. \ref{confusion}.

\textbf{A2-Mesh prediction coherence from our supervised learning.} In Fig. \ref{coherence}, we display coherence of test-time face mesh predictions from our supervised learning setting. We use different utterances from the same speaker at different time-step as inputs to produce the meshes. 
In Fig. \ref{coherence}, one can observe from, for example, jaw widths that the output meshes are different for the two speakers;
by contrast, meshes for the same speaker are highly coherent.
These results demonstrate that our training strategy successfully predicts coherent geometry for the same speaker and can predict different topologies for different identities. 
Finally, this coherence illustration also implies advantages over previous voice-to-face methods that work on the image domain \cite{NEURIPS2019_eb9fc349, oh2019speech2face}. Their generation includes variations of background, hairstyles, and the rest. In contrast, Our results exclude these variations and focus on facial geometry to validate the correlation between face shapes and voices. 

\textbf{A3+A4-Comparison against the baseline and the major improvement from our supervised learning.} We further compare against Base-2 (See Sec.4-Baseline in the paper: the cascaded pretrained blocks). One reference face, one 3D face mesh produced by our method, and one by Base-2 are presented in Fig. \ref{supervised_comp}.
For the example on the left, the person of interest has a wider jawbone, and the mesh produced from our method also shows a similar trait. On the right, the image shows a wider face shape and apparent cheeks. Our 3D model also displays a similar shape, but Base-2 shows a much thinner face. Our mesh can reflect the wideness of faces, which corresponds to our findings in paper Table 1 that the major improvement voice can hint is ER (ear-to-ear ratio).  
In summary, we use the above visual results to show that the supervised learning of the analysis framework is effective.
Fig. \ref{supervised_gt} shows the output face meshes have similar overall face shapes to the reference images, which shows the model's ability for various types of faces and is validated in Fig. \ref{confusion}.
Fig. \ref{coherence} shows our supervised method can predict coherent face shapes.
Fig. \ref{supervised_comp} shows the output face models are more similar to the reference than Base-2 in terms of overall face shapes, which again validates the ER improvements shown in the paper Table 1.

\begin{figure}[htb!]
\begin{center}
\includegraphics[width=0.8\linewidth]{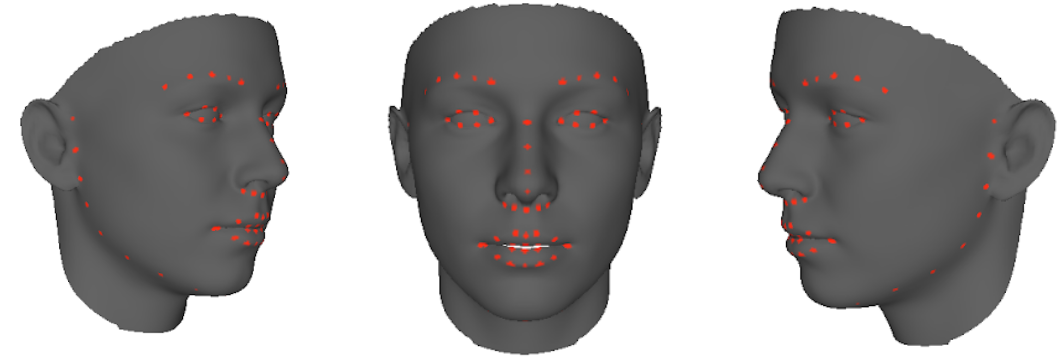}
\end{center}
  \vspace{-18pt}
  \caption{\textbf{Illustration of commonly-used 68-point 3D facial landmarks.}}
\label{landmark}
\end{figure}

\begin{figure}[bt!]
\begin{center}
\includegraphics[width=0.50\linewidth]{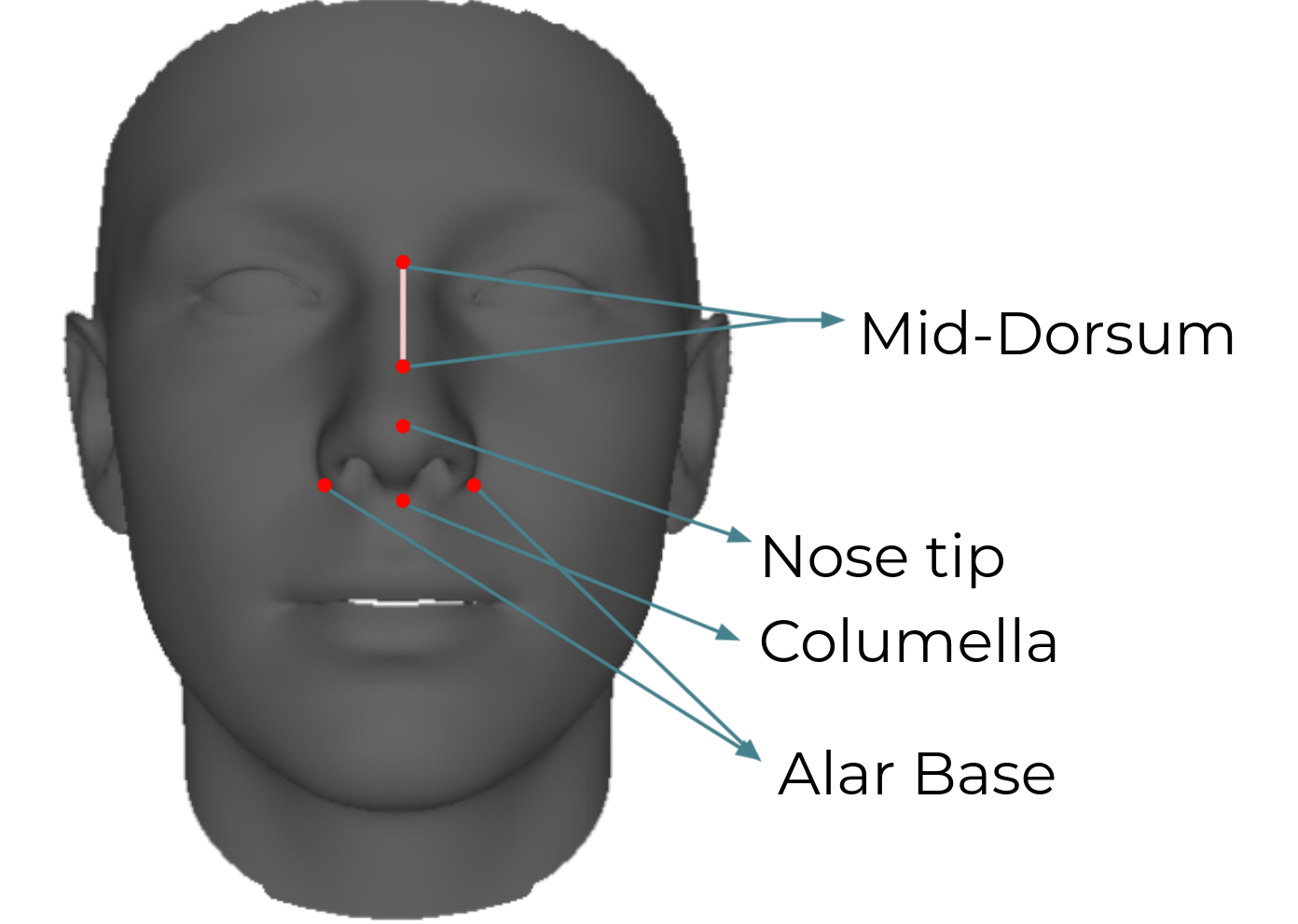}
\end{center}
  \vspace{-15pt}
  \caption{\textbf{Illustration of physiology terms in Sec. \ref{sec:point-based}}. }
\label{terms}
%\vspace{-1pt}
\end{figure} 

\clearpage

\begin{figure*}[bt!]
\begin{center}
\includegraphics[width=0.75\linewidth]{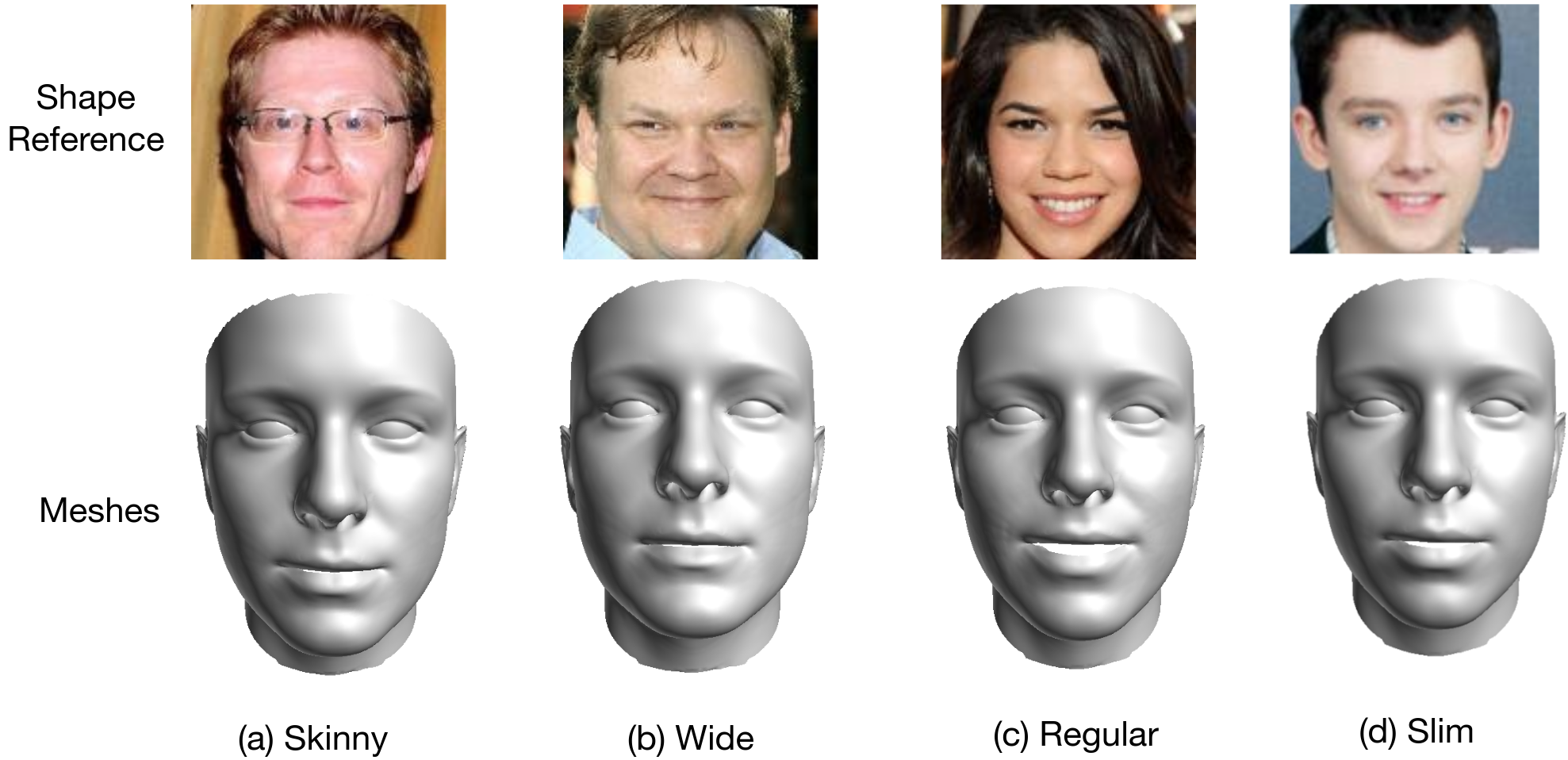}
\end{center}
  \vspace{-12pt}
  \caption{\textbf{Visualization of predicted 3D face meshes from our supervised learning}. We display four face shapes, skinny, wide, regular, and slim, and their reference images to show the shape correspondence. References are provided to identify face shapes of the person of interest.}
\label{supervised_gt}
\vspace{-15pt}
\end{figure*}

\begin{figure*}[bt!]
\begin{center}
\includegraphics[width=0.75\linewidth]{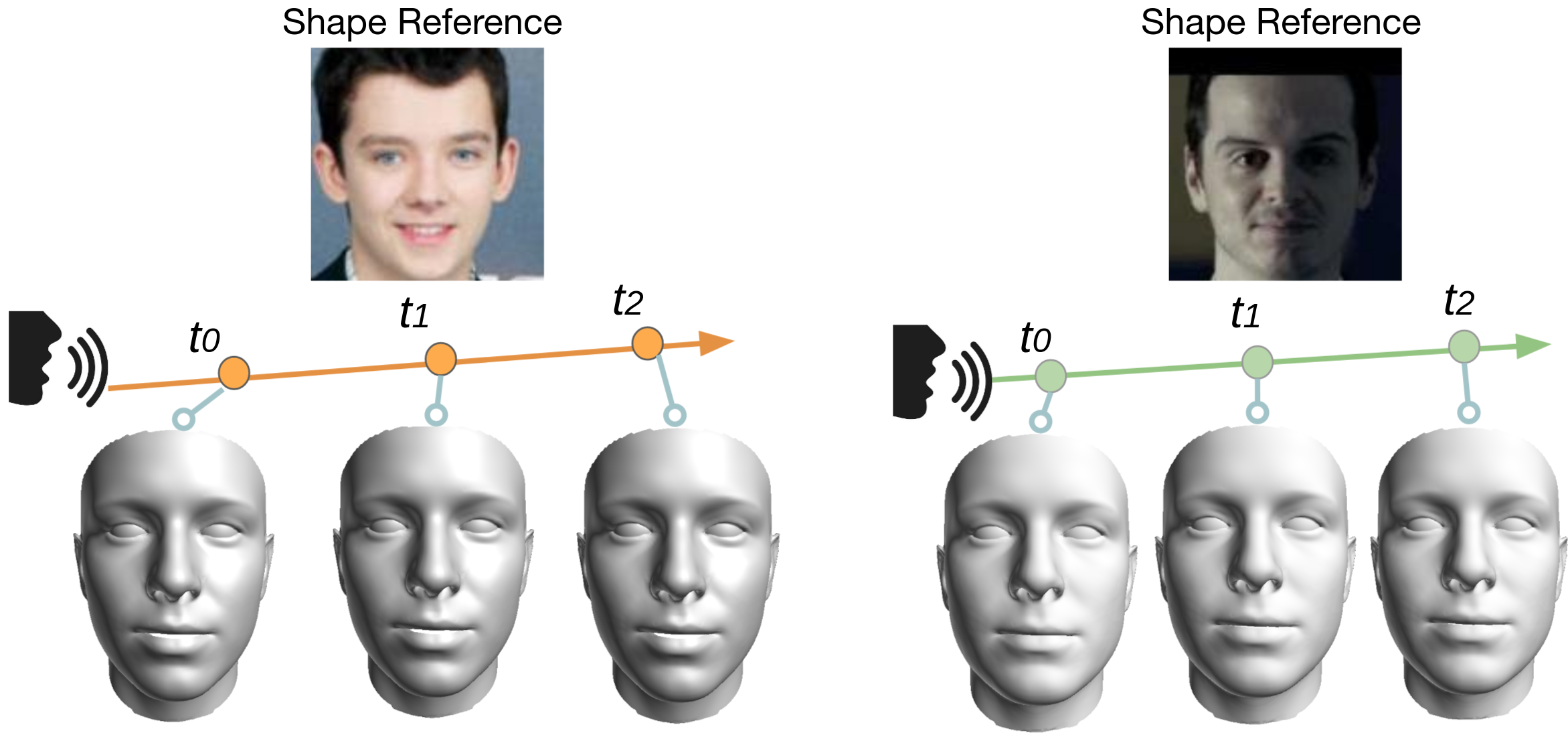}
\end{center}
  \vspace{-8pt}
  \caption{\textbf{Inference coherence of meshes produced from our CMP- supervised learning.} }
\label{coherence}
\vspace{-6pt}
\end{figure*} 

\begin{figure*}[bt!]
\begin{center}
\includegraphics[width=0.70\linewidth]{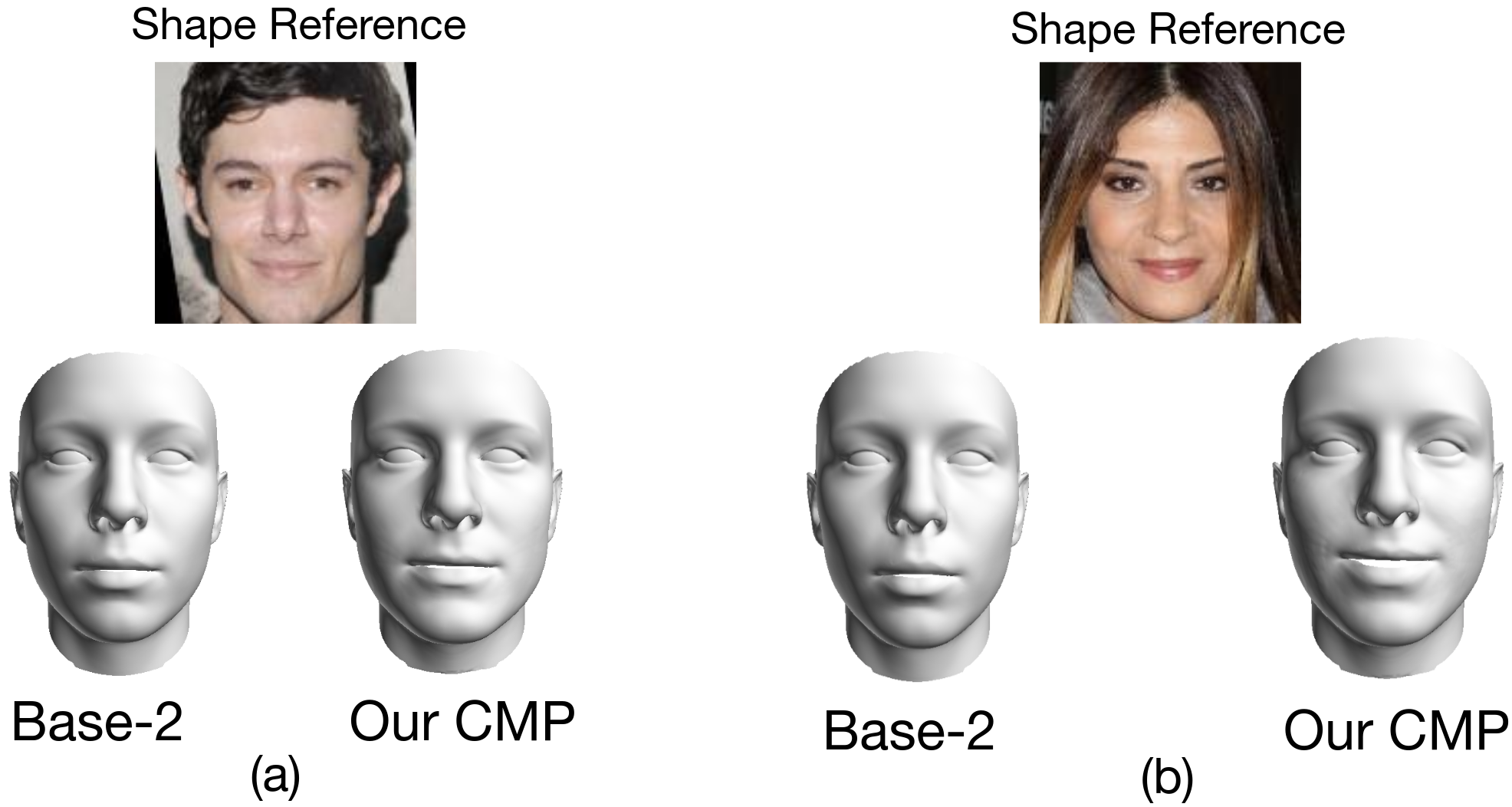}
\end{center}
  \vspace{-16pt}
  \caption{\textbf{Comparison of output face meshes from our CMP of the supervised learning and Base-2}. In case (a), our mesh shows a more squared face with a wider jawbone, but Base-2 only shows a slim face. Reference face in (b) is wider and bears apparent cheeks, and our result is much more similar to the reference.}
\label{supervised_comp}
\vspace{-7pt}
\end{figure*} 

\clearpage

\section{Point-based and Region-based Metrics}
\label{sec:point-based}
Normalized Mean Error (\textbf{NME}, point-based) of facial landmarks. BFM Face annotates 68 points 3D facial landmark points that lie on the eyes, nose, mouth, and face outlines (shown in Fig. \ref{landmark}). We calculate the NME of the landmark point set between the predicted and reference 3D face meshes, i.e., first calculate the Euclidean distance of two landmark sets and then normalize the distance by the face size (square root of face width $\times$ length). 

Results in Table~\ref{NME_metric} show NME for 3D facial landmark alignment. Quantitative results under this metric show improvements, but the gains are smaller. It is because most facial landmarks concentrate on eyes, nose, and mouth parts that naturally bear more minor deformations across people. For example, the nose tip and mid-dorsum usually lie on the centerline of faces, and alar base and columella are located around them closely. (See Fig. \ref{terms} for the definition of these physiology terms.)   

\begin{figure}[hbt!]
\begin{center}
\includegraphics[width=0.3\linewidth]{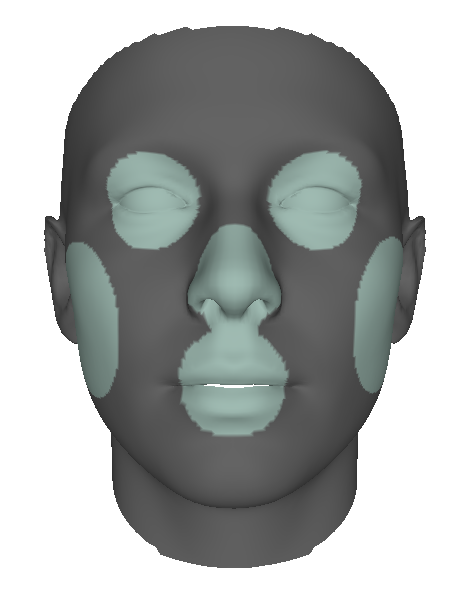}
\end{center}
  \vspace{-14pt}
  \caption{\textbf{Illustration of regions.} We show regions of the left eye, right eye, nose, mouth, left cheek, and right cheek that are used in the part-based point-to-plane evaluation using ICP in our paper Table 3.}
\label{regions}
\vspace{-5pt}
\end{figure}

Point-to-Plane Root Mean Square Error (\textbf{Point-to-Plane RMSE}, region-based). We follow the surface registration for 3D models using the popular iterative closest point (ICP) \cite{pomerleau2015review} algorithm to align the predicted and reference meshes. We then calculate point-to-plane RMSE. Registration for the holistic face and facial parts (illustrated in Fig.~\ref{regions}) are considered and shown in Table~\ref{P2PRMSE_metric} and Table~\ref{part_metric}. Both supervised and unsupervised CMP outperforms the baselines in either holistic or part-based registrations. These evaluations indicate the capability of cross-modal learning, from voice inputs to 3D face outputs.

\begin{table}[tb!]
\begin{center}
  \caption{\textbf{NME for point-based metric study.} 68 facial landmarks annotated in BFM Face \cite{paysan20093d} are used for measurements.}
  \vspace{-7pt}
  \small
  \label{NME_metric}
  \begin{tabular}[c]
  {|
  p{1.5cm}<{\centering\arraybackslash}|
  p{0.9cm}<{\centering\arraybackslash}|
  p{0.9cm}<{\centering\arraybackslash}|
  p{1.4cm}<{\centering\arraybackslash}|
  p{1.45cm}<{\centering\arraybackslash}|}
  \hline
      Landmark Alignment  & Base-1 & Base-2 & CMP- supervised & CMP- unsupervised  \\
    \hline
       NME & 0.2979 & 0.2969 & 0.2723 & 0.2904 \\
    \hline
  \end{tabular}
  \vspace{-18pt}
\end{center}
\end{table}

\begin{table}[tb!]
\begin{center}
  \caption{\textbf{Point-to-Plane RMSE study.} ICP is used to align the predicted and reference meshes. We calculate point-to-plane RMSE after ICP.}
  \vspace{-12pt}
  \label{P2PRMSE_metric}
  \small
  \begin{tabular}[c]
  {|
  p{1.6cm}<{\centering\arraybackslash}|
  p{0.9cm}<{\centering\arraybackslash}|
  p{0.9cm}<{\centering\arraybackslash}|
  p{1.4cm}<{\centering\arraybackslash}|
  p{1.45cm}<{\centering\arraybackslash}|}
  \hline
      Holistic Registration & Base-1  & Base-2 & CMP- supervised & CMP- unsupervised  \\
    \hline
      RMSE & 1.357 & 1.348 & 1.210 & 1.312 \\
    \hline
  \end{tabular}
  \vspace{-10pt}
\end{center}
\end{table}

\begin{table}[tb!]
\begin{center}
  \caption{\textbf{Part-based point-to-plane RMSE study.} 
   %We define six common facial parts: left eye, right eye, nose, mouth, left cheek, and right cheek. We perform ICP for these parts and calculate the point-to-plane RMSE.
  }
  \label{part_metric}
  \vspace{-10pt}
  \small
  \begin{tabular}[c]
  {|
  p{1.7cm}<{\centering\arraybackslash}|
  p{0.9cm}<{\centering\arraybackslash}|
  p{0.9cm}<{\centering\arraybackslash}|
  p{1.35cm}<{\centering\arraybackslash}|
  p{1.4cm}<{\centering\arraybackslash}|}
  \hline
       Part Registration  & Base-1 & Base-2 & CMP- supervised & CMP- unsupervised  \\
    \hline
       Left Eye & 0.3961 & 0.3945 & \textbf{0.3517} & 0.3779 \\
       Right Eye & 0.3667 & 0.3656 & \textbf{0.3349} & 0.3488 \\
       Nose & 0.5258 & 0.5250 & \textbf{0.5141} & 0.5177 \\
       Mouth & 0.3466 & 0.3435 & \textbf{0.2958} & 0.3149 \\
       Left Cheek & 0.4748 & 0.4735 & \textbf{0.4654} & 0.4711 \\
       Right Cheek & 0.5078 & 0.5061 & \textbf{0.4916} & 0.4919 \\
    \hline
  \end{tabular}
  \vspace{-18pt}
\end{center}
\end{table}

\section{Simple Oracles}
\label{sec:oracles}
We provide numerical results of simple oracles as other baselines. Oracle (1): We take average 3D faces in the Voxceleb-3D training set and use the mean shape directly as predictions for testing data; Oracle (2): We take the mean 3D face for the male/female group in the training set and use the mean shape for male/female as predictions at test time. The following Table \ref{tab:oracles} shows results using line-based (Mean ARE), point-based (NME), and region-based (RMSE) metrics, compared with Base-2 described in paper paper Sec.4-Baseline. Simple oracles perform worse than Base-2, which means directly taking average faces is naive and weaker than the network-based solutions. This validates our baseline construction, and our CMP framework can further predict more accurate face geometry from voices for each person of interest. 

\begin{table*}[htb]
\begin{center}
  \caption{\textbf{Study on performances of different KD strategies.} ARE is used as the metric for comparison.}
  \vspace{-8pt}
  \label{KD_study}
  \footnotesize
  \begin{tabular}[c]
  {|
  p{1.20cm}<{\centering\arraybackslash}|
  p{1.30cm}<{\centering\arraybackslash}|
  p{1.30cm}<{\centering\arraybackslash}|
  p{1.30cm}<{\centering\arraybackslash}|
  p{1.50cm}<{\centering\arraybackslash}|
  p{1.30cm}<{\centering\arraybackslash}|
  p{1.30cm}<{\centering\arraybackslash}|
  p{1.30cm}<{\centering\arraybackslash}|
  p{1.30cm}<{\centering\arraybackslash}|}
  \hline
      ARE  & Vanilla KD & Attention & SP & Correlation & RKD & CRD & VID & PKT\\
    \hline
       ER & 0.0306 & 0.0318 & 0.0230 & 0.0227 & 0.0172 & 0.0198 & 0.0213 & 0.0184\\ 
       FR & 0.0173 & 0.0172 & 0.0169 & 0.0173 & 0.0171 & 0.0172 & 0.0172 & 0.0172\\ 
       MR& 0.0173 & 0.0173 & 0.0179 & 0.0179 & 0.0195 & 0.0177 & 0.0178 & 0.0176\\ 
       CR & 0.0540 & 0.0551 & 0.0471 & 0.0471 & 0.0474 & 0.0481 & 0.0471 & 0.0484\\
       \hline
           Mean & 0.0298 & 0.0304 & 0.0262 & 0.0263 & \textbf{0.0254} & 0.0255 & 0.0259 & \textbf{0.0254}\\
    \hline
  \end{tabular}
  %\vspace{-17pt}
\end{center}
\end{table*}

\begin{table}[ht]
\caption{\textbf{Oracle results}. We show quantitative evaluations of simple oracles explained in Sec. \ref{sec:oracles}.}
\begin{center}
  \vspace{-18pt}
  \footnotesize
  \begin{tabular}[c]
  {|
  p{1.9cm}<{\centering\arraybackslash}|
  p{1.0cm}<{\centering\arraybackslash}|
  p{1.2cm}<{\centering\arraybackslash}|
  p{1.2cm}<{\centering\arraybackslash}|
  p{1.25cm}<{\centering\arraybackslash}|}
  \hline
       Metrics  & Type & Oracle(1) & Oracle(2) & Base-2  \\
    \hline
       Mean ARE & line & 0.0319 & 0.0311 & 0.0302 \\
       NME & point & 0.3058 & 0.3021 & 0.2969\\
       RMSE & region & 1.540 & 1.529 & 1.348 \\
    \hline
  \end{tabular}
  \vspace{-19pt}
  \label{tab:oracles}
\end{center}
\end{table}

\section{Pose from the Pretrained Expert}
\label{sec:pose}
Here we study the robustness of head poses estimated from the expert, which helps our visualization (in Fig.7-10 of the paper) of laying 3D face meshes onto images to show the fitness.
SynergyNet \cite{wu2021synergy} as an expert used in the unsupervised framework predicts 3DMM parameters ($\alpha_s$ and $\alpha_e$) as pseudo-groundtruth based on images synthesized from GAN. Here we verify the robustness of pose estimation from the expert. As illustrated in our paper Fig. 8, synthesized faces from GAN are almost frontal because face images in VGGFace \cite{BMVC2015_41} for GAN-training are with small poses. We adopt widely-used AFLW2000-3D \cite{zhu2016face} including 2K in-the-wild face images to examine the performance of head pose estimation. Then, we calculate the mean absolute error (MAE) for three estimated Euler angles (yaw, pitch, roll). MAE is 1.49 degrees for faces whose yaw angle (left/right turns) lies in [-15$^{\circ}$, 15$^{\circ}$]. This result justifies the robustness of head pose estimation from the pretrained expert.

\section{Ablation Study on Knowledge Distillation}
\label{sec:study_unsupervised}

We conduct an extensive survey for the performance of various recent KD strategies on our unsupervised framework. 
We include vanilla KD \cite{hinton2015distilling}, Attention \cite{zagoruyko2016paying}, SP \cite{tung2019similarity}, Correlation \cite{peng2019correlation}, RKD \cite{park2019relational}, CRD \cite{tian2019crd}, VID \cite{ahn2019variational}, PKT \cite{passalis2020probabilistic}, and train our unsupervised framework with different $\mathcal{L}_{\textit{KD}}$. 
Here we show the results in Table \ref{KD_study}. 
We find that more recent and advanced KD methods attain similar results. For example, RKD, CRD, and PKT have very close performance, compared with earlier methods such vanilla version or using attention map similarity. Therefore, the study validates our adoption of conditional probability in our paper Eq.(4)\footnote{The scaled and shifted cosine similarity is  $K_{cosine}(z_i,z_j)=\frac{1}{2} \left((z_i^E z_j / \|z_i^E\|_2 \|z_j\|_2) +1 \right).$}, introduced in PKT \cite{passalis2020probabilistic}.

\begin{figure}[bt!]
\begin{center}
\includegraphics[width=1.0\linewidth]{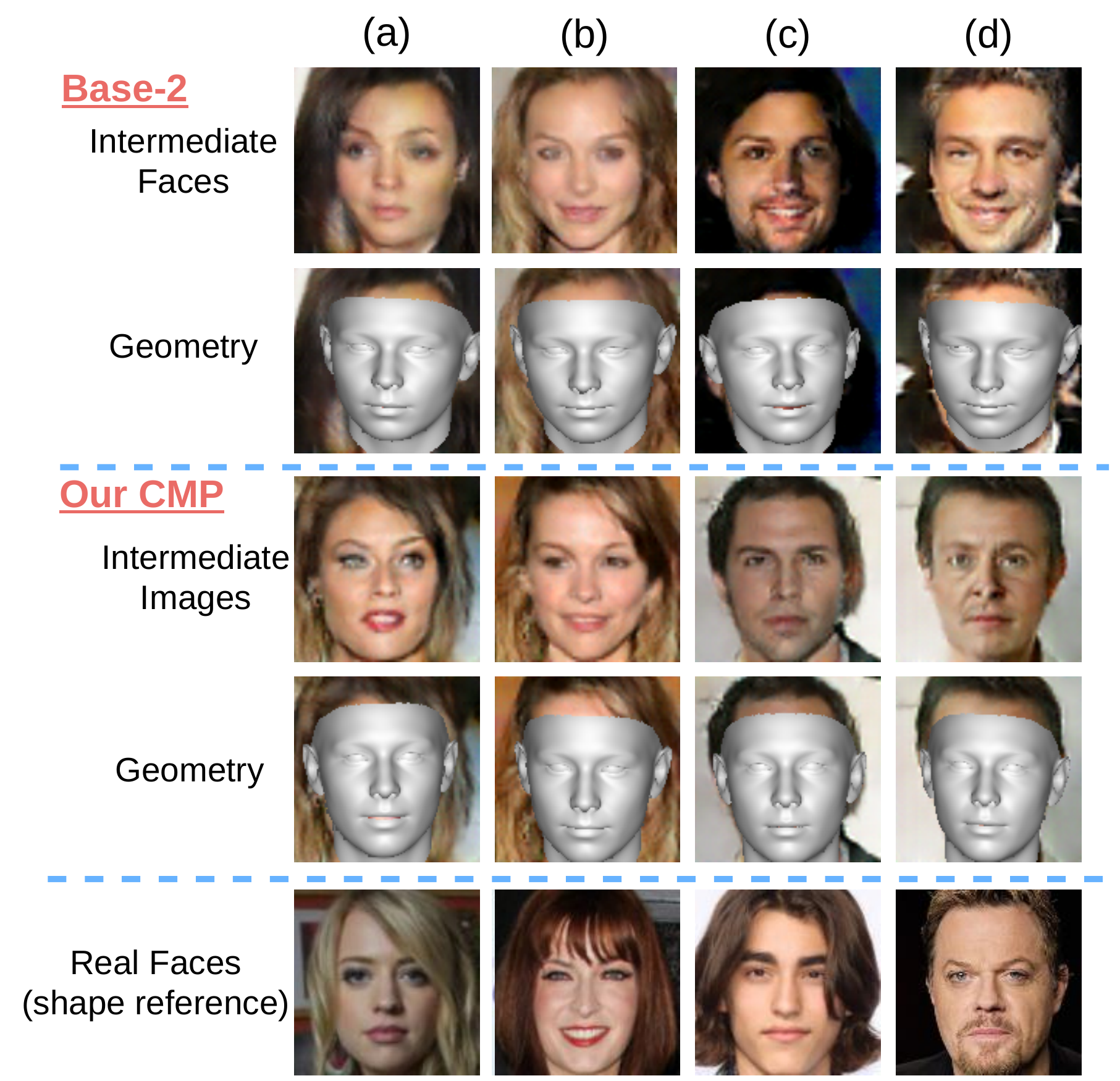}
\end{center}
  \vspace{-15pt}
  \caption{\textbf{Qualitative comparison between our unsupervised CMP and the baseline}. This figure presents more results that extend Fig. 10 of the main paper.}
\label{unsupervised_comp_supp}
\vspace{-12pt}
\end{figure}

Further, we exhibit more qualitative comparisons in Fig. \ref{unsupervised_comp_supp} extending Fig. 10 of the main paper.

% \section{Illustration of Regions}
% \label{sec:facialPoints}
% In Fig. \ref{regions}, we show the areas of facial parts used in our paper Sec. 4.2 and Table 3 for local region registration using ICP.

% \begin{figure}[hbt!]
% \begin{center}
% \includegraphics[width=0.45\linewidth]{Figures/regions.png}
% \end{center}
%   %\vspace{-14pt}
%   \caption{\textbf{Illustration of regions.} We show regions of the left eye, right eye, nose, mouth, left cheek, and right cheek that are used in the part-based point-to-plane evaluation using ICP in our paper Table 3.}
% \label{regions}
% \vspace{-5pt}
% \end{figure} 

\section{Applications of Voice-to-3D-Face Task}
\label{sec:applications}
We focus on the analysis that purposes to validate the correlation between voice and 3D face geometry. Here we describe more on application sides, where our work has potential at:

1. Our work can be used for public security, such as recovering the face shape of the unheard speech of a suspect or a masked robber.

2. Our work can generate personal avatars in gaming or virtual reality systems: it is helpful to create a rough 3D face model from voices as the initialization, and users can refine its shape based on one's preference.

3. 3D faces from voice can potentially provide another verification mode for person identification other than speech and face image verification. 

\section{Limitations}
\label{sec:limitation}
Our work focus on the analysis between voices and 3D faces, and generating high-quality meshes is not our aim. In fact, using voices as inputs to produce face meshes has its inherent limitations since our face wideness might be gleaned from voices from our intuition, but more subtle details, such as bumps or wrinkles, cannot be hinted at from this modality.
We target at analysis between one's normal voices and face shapes and utilize Voxceleb as the speech source, which is primarily interviews or talks. As pointed out in main paper Sec.5- Ethical statement, more implicit factors like talking after drinking or abnormal health conditions may affect the analysis, but this requires large data corpse from a medical or physiological view to further validate these effects. 

% \section{Computing Infrastructure}
% \label{sec:infra}
% We run experiments on a machine with Intel Core i9-7900X CPU, Nvidia GTX-1080Ti GPU, 64GB RAM, Linux 16.04 OS. We implement and test our system on Python 3.8 and PyTorch v1.7.1.

\end{document}